%% file: main.tex
\documentclass{article}

    \PassOptionsToPackage{numbers, compress}{natbib}

    \usepackage[preprint]{neurips_2025}

\usepackage[utf8]{inputenc} % allow utf-8 input
\usepackage[T1]{fontenc}    % use 8-bit T1 fonts
\usepackage{hyperref}       % hyperlinks
\usepackage{url}            % simple URL typesetting
\usepackage{booktabs}       % professional-quality tables
\usepackage{amsfonts}       % blackboard math symbols
\usepackage{nicefrac}       % compact symbols for 1/2, etc.
\usepackage{microtype}      % microtypography
\usepackage[table]{xcolor}   % <—— 记得这个，否则 \columncolor 也会报 undefined
\usepackage{xspace}
\usepackage{colortbl}  
\usepackage{subcaption}
\usepackage{graphicx} 
\usepackage{pifont} 
\usepackage{afterpage}
\usepackage{arydshln}

\usepackage{enumitem}

\newcommand{\ours}{HumanPCR\xspace}
\newcommand{\LP}{Human-P\xspace}
\newcommand{\LC}{Human-C\xspace}
\newcommand{\LR}{Human-R\xspace}

\usepackage{float}
\usepackage{multirow}
\usepackage{changepage} 
\definecolor{mygray}{gray}{0.6} % 定义灰色

\title{HumanPCR: Probing %Perception, Comprehension, and Reasoning of 
MLLM Capabilities \\in Diverse Human-Centric Scenes}

\author{%
  \small \textbf{Keliang Li}$^{1,2}$\thanks{Equal contribution. Author order was determined randomly.}\,, 
  \textbf{Hongze Shen}$^{1,2,3}$\footnotemark[1]\,, 
  \textbf{Hao Shi}$^{1,2}$, 
  \textbf{Ruibing Hou}$^{1}$\thanks{Corresponding author.}\,, 
  \textbf{Hong Chang}$^{1,2}$, \\
  \textbf{Jie Huang}$^{1,2}$, 
  \textbf{Chenghao Jia}$^{1,2}$,
  \textbf{Wen Wang}$^{1,2}$, 
  \small \textbf{Yiling Wu}$^{3}$, \\
  \textbf{Dongmei Jiang}$^{3}$, 
  \textbf{Shiguang Shan}$^{1,2}$, 
  \textbf{Xilin Chen}$^{1,2}$ \\ 
  \small {$^1$Key Laboratory of Intelligent Information Processing of Chinese Academy of Sciences (CAS),} \\ 
  \small {Institute of Computing Technology, CAS, China} \\ 
  \small {$^2$University of Chinese Academy of Sciences, China}, \small {$^3$Peng Cheng Laboratory, China} \\
}

\begin{document}

\maketitle

\begin{abstract}\label{sec:abstract}
The aspiration for artifical general intelligence, fueled by the rapid progress of multimodal models, demands human-comparable performance across diverse environments.
We propose \textbf{\ours}, an evaluation suite for probing MLLMs’ capacity about human-related visual contexts % performance?
across three hierarchical levels: \textbf{P}erception, \textbf{C}omprehension, and \textbf{R}easoning (denoted by \LP, \LC, and \LR, respectively).
\LP and \LC feature over 6,000 human-verified multiple-choice questions, assessing massive tasks of 9 dimensions, including but not limited to essential skills frequently overlooked by existing benchmarks.
\LR offers a challenging manually curated video reasoning test that requires integrating multiple visual evidences, proactively extracting context beyond question cues, and applying human-like expertise. Each question includes human-annotated Chain-of-Thought (CoT) rationales with key visual evidence to support further research.
Extensive evaluations on over 30 state-of-the-art models exhibit significant challenges in human-centric visual understanding, particularly in tasks involving detailed space perception, temporal understanding, and mind modeling. 
Moreover, analysis of \LR reveals the struggle of models in extracting essential proactive visual evidence from diverse human scenes and their faulty reliance on query-guided retrieval. 
Even with advanced techniques like scaling visual contexts and test-time thinking yield only limited benefits. We hope \ours and our findings will advance the development, evaluation, and human-centric application of multimodal models.
\end{abstract}

\afterpage{
\begin{figure*}[htbp]
  \centering
  \begin{minipage}[b]{0.345\textwidth}
    \includegraphics[width=\linewidth]{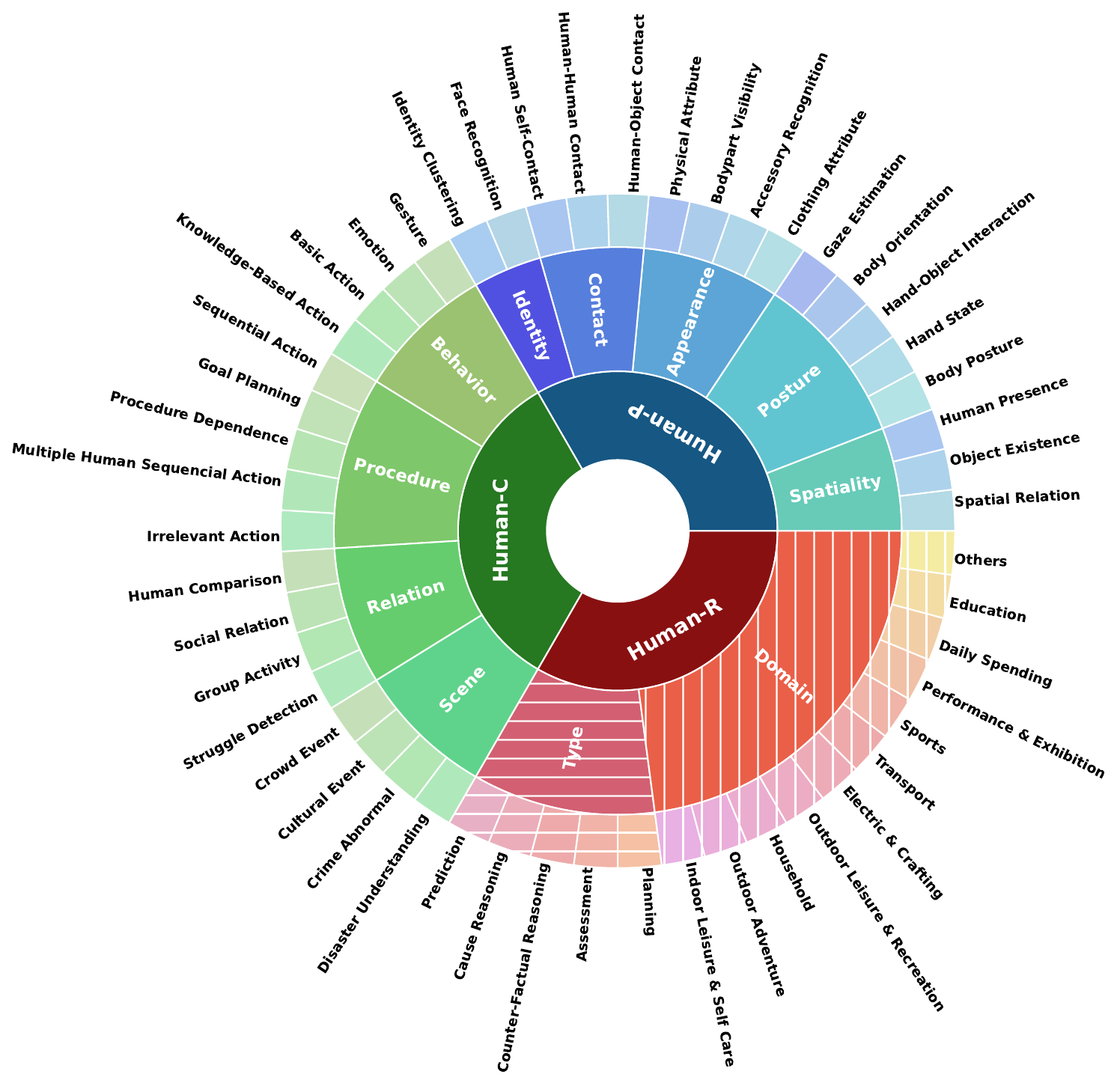}
    \vspace{0pt} % 左上/左下之间的间距；设为 0pt 就真正无缝
    \includegraphics[width=\linewidth]{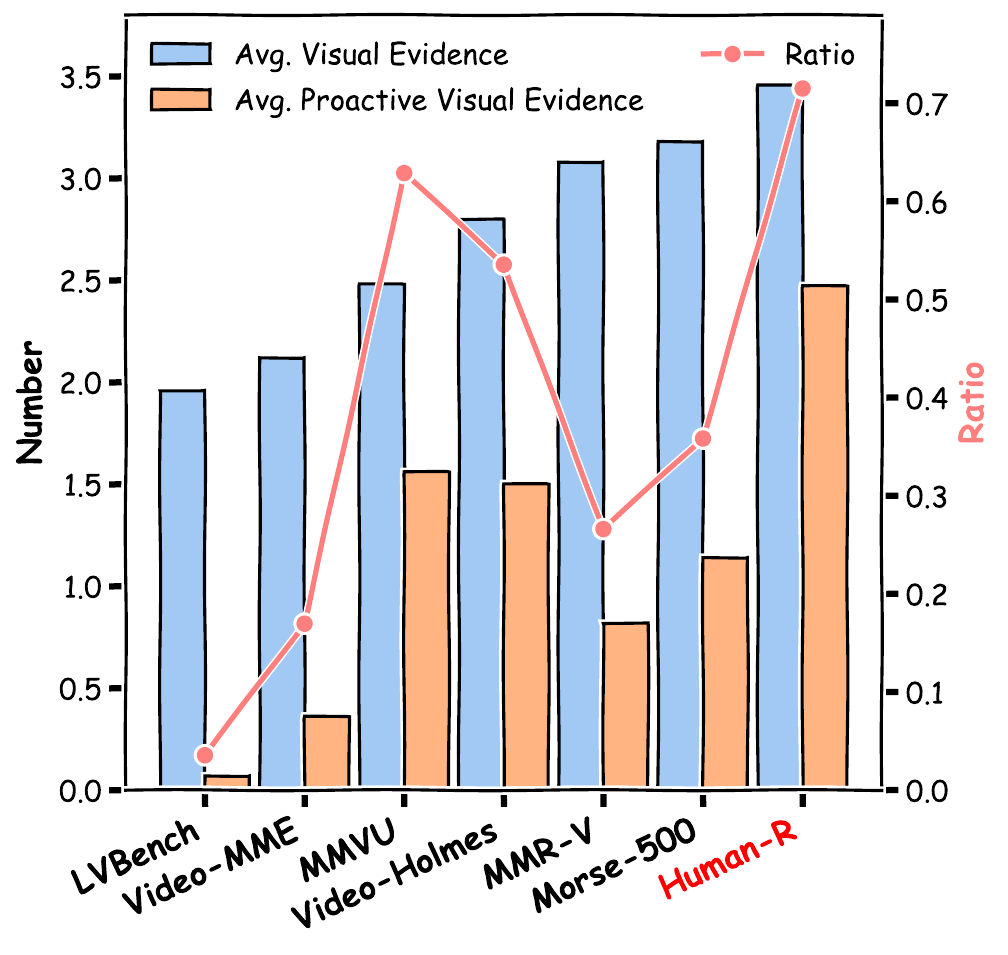}
  \end{minipage}%
  \hspace{0pt}                 % 调整左右间隔（可改成 \hspace{2pt} 等）
  \begin{minipage}[b]{0.64\textwidth}
    \includegraphics[width=\linewidth]{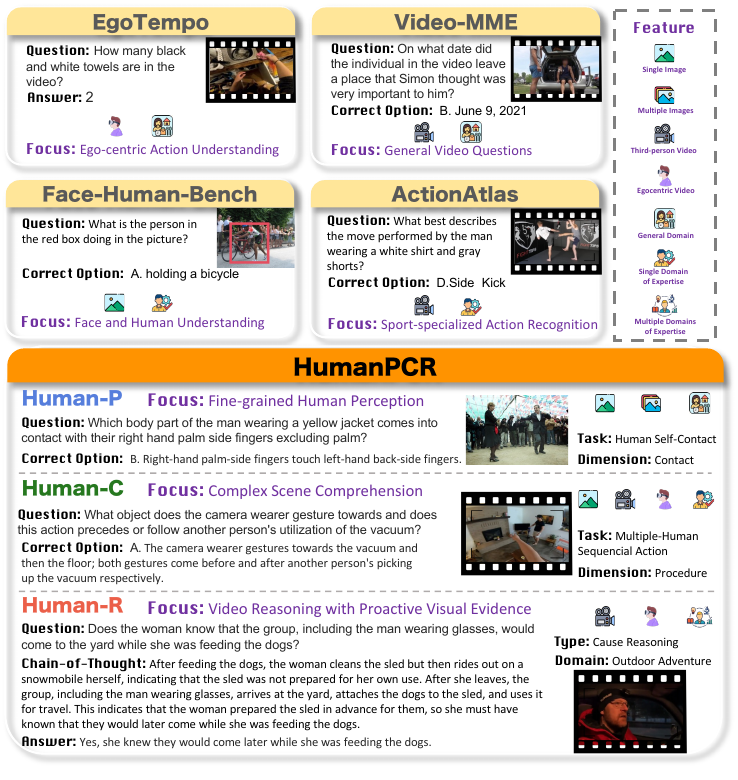}
  \end{minipage}
  \caption{{Overview of \ours.} 
(Left–upper) The hierarchical taxonomy of HumanPCR. 
(Left–lower) Quantitative comparison of \emph{visual evidence} across representative benchmarks and \ours, showing the markedly higher proactive-reasoning demand of our benchmark. 
(Right) Qualitative comparisons between existing benchmarks and \ours.
}
  \label{fig:overview}
\end{figure*}
}

\section{Introduction} \label{sec:intro}
The rapid advancement of Multimodal Large Language Models (MLLMs) has shown remarkable potential in understanding diverse contexts \cite{llava_ov,llavavideo,gemini,gpt-4o,qwen2_5,qwenvl2}. This progress fuels the aspiration toward artificial general intelligence and raises the expectation for MLLMs to operate effectively in the wild like humans \cite{ceval,mmlu,mmmu}. A key prerequisite lies in the ability to understand humans in diverse, complex, and dynamic contexts, as human behavior inherently reflects the complexities of the world as well as real-world interaction 
\cite{ego4d,egoexo4d,jrdb,social_jrdb,activitynet}. In this work, we take this perspective and systematically investigate how well MLLMs understand humans across varied scenes, focusing on the critical aspects of perception, comprehension, and reasoning in human-centric visual understanding.

Human-centric visual understanding \cite{humanbench,unihcp} remains a fundamental challenge in artificial intelligence, specially as recent benchmarks~\cite{motionbench,face_human,zhou2024humanvbench,egoschema,actionatlas,feng2023posegpt,yollava} tend to focus on specific tasks and scenarios (as in examples of Figure~\ref{fig:overview}, top right). While these specialized datasets advance our knowledge in subdomains like ego-centric action or face understanding, they are inherently limited in scope and fail to capture the broad complexity of human activities. Meanwhile, general MLLM benchmarks \cite{mmmu,videomme,mmbench,mmtbench,zhou2024mlvu} cover a wider range of topics but often lack the depth needed for fine-grained human-centric evaluation, overlooking intricate tasks such as gaze or body orientation estimation \cite{gaze360,posescript}. With the rapid advancement of reasoning-specialized models~\cite{GoogleDeepMind2025Gemini25,OpenAI2025o3o4mini,deepseek-r1}, recent benchmarks~\cite{mmmu,mmvu,mmrv,videommmu,m3cot,videomme,lu2023mathvista} have sought to increase evaluation complexity to better test models’ ability to tackle real-world challenges, but they still fall short in visual-centric complexity. As shown in Figure \ref{fig:overview} (the lower left bar chart) and Figure \ref{fig:examples}, current benchmarks~\cite{lvbench,morse500,videoholmes} often require less visual evidence and rarely assess models’ ability to proactively seek and reason with implicit visual cues beyond what is stated in the question. In contrast, the dense and dynamic human activities found in real-world settings—which are rich sources of intelligence \cite{worldqa,egonormia,mmtom}—remain largely overlooked, as existing benchmarks seldom challenge models’ reasoning in these truly complex, human-centered scenarios \cite{mmvu,videommmu,qi2025vcr}.

To address the aforementioned gaps, this work contributes an evaluation suite, \textbf{\ours}, for benchmarking human-centric visual understanding. \ours is structured into a hierarchical taxonomy of three levels: \textbf{\LP} for perception, \textbf{\LC} for comprehension, and \textbf{\LR} for reasoning, as detailed in the upper left of Figure \ref{fig:overview}. 
For \LP and \LC, a large-scale multiple-choice question-answering (QA) dataset is curated, featuring over 6,000 human-verified image- and video-based QA pairs, which assesses perception and comprehension capabilities along 9 dimensions, encompassing a total of 34 distinct tasks. \LR presents a challenging, manually curated, open-ended video reasoning dataset. Sourced from 11 diverse human-related domains, Human-R requires models to solve complex problems by integrating multiple visual evidences, proactively extracting context beyond explicit cues in the questions, and applying human-like domain expertise. 
To support further research, each  question in \LR is augmented with human-annotated Chain-of-Thought (CoT) \cite{cot} rationales covering all key visual evidence. 

We comprehensively evaluate a large suite of open-source and proprietary models on \ours. Our findings reveal significant challenges for existing models in human-centric visual understanding and capability limitations in tasks involving detailed spatial perception \cite{thinking_in_space}, temporal understanding \cite{videomme,zhou2024mlvu}, and mind modeling \cite{ego4d,egonormia,mmtom}.
Furthermore, analysis of \LR highlights models' struggles in reasoning on multiple visual evidence from diverse human scenes and their faulty reliance on query-guided cues \cite{longvideobench,video-rag,longvu}. Consequently, merely scaling visual contexts \cite{gemini,gpt-4o,qwen2_5,adaretake,longvu} or increasing test-time thinking \cite{OpenAI2025o3o4mini,internvl3,self-refine} yields only limited benefits. Missing proactive evidence and recognition errors account for a substantial proportion of wrong cases in \LR. 
These results validate the necessity of our benchmark and imply critical shortcomings for MLLMs in solving real-world problems. We hope this benchmark will help to adapt MLLMs to diverse downstream applications and advance the development of more general and capable MLLMs.

\definecolor{proactiveyellow}{RGB}{250,231,160}
\definecolor{referredblue}{RGB}{194,213,236}
\begin{figure*}[tbp]
    \centering
    \includegraphics[width=1\textwidth]{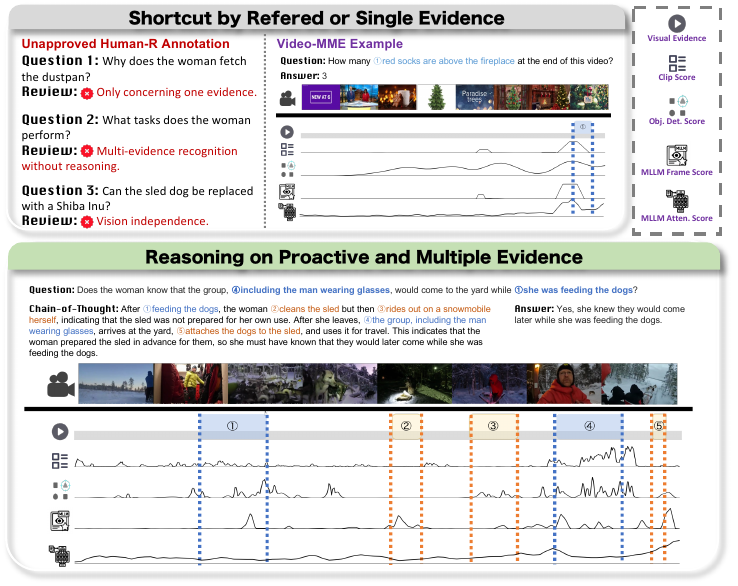}
    \caption{Illustration of \LR and Video-MME examples. The distribution of {\setlength\fboxsep{1pt}\colorbox{proactiveyellow}{proactive}} and {\setlength\fboxsep{1pt}\colorbox{referredblue}{referred}} \textit{visual evidence} is highlighted, while the score of recent video context extraction methods based on (1) Clip Score, (2) Object Detection Score, (3) MLLM Single-Frame Score, and (4) MLLM Attention Score, are also visualized. (Upper) The referred and sparse evidence required enables shortcuts without comprehensive video reasoning, so we exclude annotations solvable by single evidence or those not requiring reasoning. (Lower) on \LR, high scores on referred/irrelevant frames, ignoring proactive evidence validate the shortcut, highlighting the necessity of multiple and proactive evidence for video reasoning evaluation.}
    \vspace{-0.5\baselineskip}
    \label{fig:examples}
\end{figure*}
\section{Related Work}
\textbf{Multimodal Large Language Models (MLLMs).} %originated from early efforts to integrate images \cite{} into 
MLLMs have evolved from Large Language Models (LLMs) to process diverse modalities, including image sequences \cite{llava,llava_ov}, video \cite{internvl,llava_ov,videollava}, and audio \cite{borsos2023audiolm}. 
Recent models like Qwen2.5-VL \cite{qwen2_5} and InternVL3 \cite{internvl3} offers dynamic resolution processing and support extremely long contexts. For long video understanding, especially within limited context windows, context extraction strategies\cite{liu2025ola,yang2024pvc} such as frame sampling \cite{aks,huang2025frag,tstar} and token pruning \cite{adaretake,zhang2025flexselect,longvu} are employed to efficiently extract relevant content \cite{video-rag,yang2024pvc}. Furthermore, the advent of models with powerful reasoning capabilities \cite{deepseek-r1,o1,GoogleDeepMind2025Gemini25,OpenAI2025o3o4mini} has spurred pioneering efforts \cite{video-r1,llava-o1,videorft,li2025videochat} to enhance visual understanding by leveraging and strengthening the model's reasoning processes.

\textbf{Evaluation on Human-Centric Visual Understanding.}
Human-centric visual understanding has long stood as a cornerstone application in artificial intelligence~\cite{humanbench,unihcp,activitynet}. Riding on the recent advances of MLLMs, a growing body of work~\cite{egonormia,egoschema,zhou2024humanvbench,li2024herm} has begun to probe their competence in decoding human motion~\cite{motionbench}, recognizing faces~\cite{face_human,mcllava,yollava}, and interpreting domain-specific activities~\cite{cui2023probio,actionatlas,plizzari2025omnia}. Yet, across studies, a recurring conclusion remains: current MLLMs still exhibit pronounced limitations in human-centric visual understanding.
Another line of work has focused on applying MLLMs to 3D human modeling tasks~\cite{feng2023posegpt,lin2025chathuman,li2025unipose}, such as motion generation~\cite{jiang2023motiongpt,chen2024motionllm} and motion tracking~\cite{hong2024egolm}.
Given the prevalence of human activities in real-world scenarios, many comprehensive benchmarks~\cite{mmbench,videomme,li2023seed} also incorporate human-centric tasks, including celebrity identification and action recognition.
However, these benchmarks often lack structured taxonomies and unified evaluation protocols, which makes it difficult to derive systematic insights into human-centric visual understanding.

\textbf{Multimodal Reasoning Benchmarks.}
With the growing capabilities of MLLMs, plenty of benchmarks have involved increasingly complex tasks to assess their reasoning abilities \cite{m3cot,mmlu,thinking_in_space,mmcode,lu2023mathvista}. 
Early studies assessed reasoning capabilities \cite{mmmu} using standardized examinations spanning multiple disciplines, and were subsequently extended to tasks involving natural images \cite{m3cot} and multi-image inputs \cite{mmiu}.
For video-based reasoning, recent progress \cite{chandrasegaran2024hourvideo,thinking_in_space,cao2025videosimple,videommlu,morse500,videoholmes,mmrv} has been made. Comprehensive benchmarks \cite{longvideobench} such as Video-MME \cite{videomme} and MMBench-Video \cite{fang2024mmbenchvideo} include predefined reasoning tasks, while MultiHop-EgoQA \cite{mh-grounded} targets reasoning over object co-occurrence.
MMWorld \cite{he2024mmworld} and MMVU \cite{mmvu} introduce multi-disciplinary video reasoning questions, while VideoMMMU \cite{videommmu} focuses on knowledge reasoning from subject explanation videos. More recent works, such as VideoEspresso \cite{han2024videoespresso} and VCR-Bench \cite{qi2025vcr}, construct video QA with reasoning processes by leveraging automatic annotations and repurposing existing QA pairs.
In contrast, our proposed Human-R distinguishes itself by providing human-annotated QA pairs along with human-annotated CoT rationales with key visual evidence, emphasizing multi-evidence integration and the challenges of information extraction under diverse human-centric contexts.

\section{The \ours Benchmark}
\label{sec:chap_3}

\subsection{Overview of \ours}

We introduce \ours benchmark to evaluate how MLLMs understand humans in real-world scenarios. \ours is structured within a three-level hierarchical taxonomy with holistic dimensions, which results in over 6,000 multiple-choice questions across 34 tasks and 442 open-ended questions includeing human-annotated CoT rationales.
More key statistics of \ours are summarized in Table~\ref{tab:key_statistics_independent}. Definitions for 
 each task, detailed implementation, and metadata of annotations are provided in Appendix~\ref{sec:appendix-taxonomy}, ~\ref{sec:appendix-data-source} and ~\ref{sec:appendix-annotation-pipeline}.

\begin{figure*}[tbp]
  \centering
  \includegraphics[width=1\textwidth]{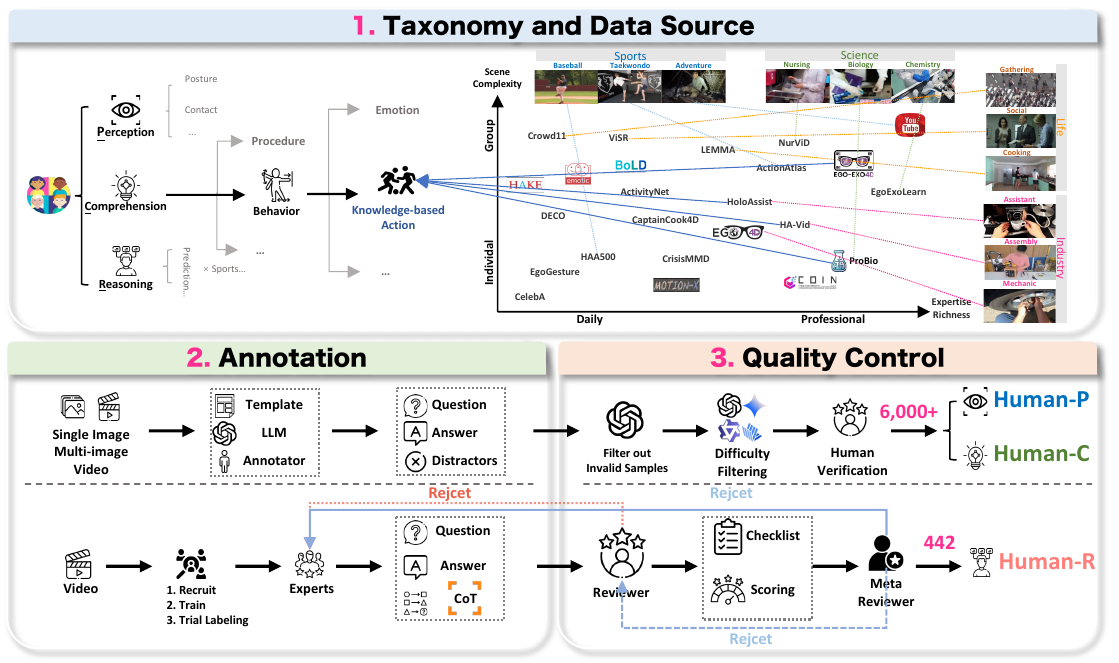}
  \caption{{Benchmark Construction.} A hierarchical task taxonomy and a diverse human-centric vision-language corpus are first built on both an extensive survey of existing literature and the requirements observed in real-world scenarios. Next, task-specific strategies including query-template and LLM repurposing, are employed to suit each subtask, and finally, all annotations are sent to a rigorous, multi-stage quality control pipeline.}
  \label{fig:construction}
\end{figure*} 

The taxonomy in \ours are briefly introduced as follows:
\vspace{-0.5\baselineskip}
\begin{itemize}[itemsep=0.3ex, topsep=0.3ex, leftmargin=2.5em]
\item \textbf{Level 1: \underline{P}erception} evaluates the recognition of specific elements. It covers five dimensions and encompasses  17 tasks: (1) \textbf{Spatiality}: perceiving existence of people, objects, and their spatial relations; (2) \textbf{Posture}: recognizing physical state and orientation of body parts, hands, and gaze; (3) \textbf{Appearance}: identifying human appearances, including inherent attributes and attirement; (4) \textbf{Contact}: recognizing detailed interaction regions between people and objects, or themselves; (5) \textbf{Identity}: recognizing a person’s identity. 
\item \textbf{Level 2: \underline{C}omprehension} assesses the comprehension of basic visual concepts integrating commonsense or domain-specific cues. It covers four dimensions and encompasses 17 tasks: (1) \textbf{Behavior}: understanding human actions and bodily movements, such as gestures and emotions; (2) \textbf{Procedure}: thoroughly understanding long-term activities, including underlying intentions and dependence among action sequences; (3) \textbf{Relation}: analyzing relations, roles and differences among individuals; (4) \textbf{Scene}: 
interpreting group dynamics or human activities within broader contexts.
\item{ \textbf{Level 3: \underline{R}easoning} examines whether models can integrate continuous, tightly coupled human dynamics within complex scenes for reasoning. We contend that the evaluation should satisfy three criteria: 
(1) \textbf{Visual Complexity}: questions should require suffient \textit{visual evidence}\footnote{In this work, \textit{visual evidence} is defined as a proposition in visual information containing single or multiple instances—such as a specific attribute of an instance or a group (e.g., orientation or action)—or as a visual relationship among multiple instances or group activities.}, and exclude redundant visual content, going beyond simple concept retrieval;
(2) \textbf{Reasoning Necessity and Diversity}: questions should engage diverse reasoning chains rather than be limited to a few reasoning patterns;
(3) \textbf{Proactivity}: questions should demand proactive extraction of visual evidence over the continuous contexts\footnote{This refers to visual information that is not, or only partially, cued by the question, termed "{\setlength\fboxsep{1pt}\colorbox{proactiveyellow}{proactive visual evidence}}", in contrast to "{\setlength\fboxsep{1pt}\colorbox{referredblue}{referred visual evidence}}" which is explicitly indicated by the question.}, rather than relying solely on evidence explicitly specified in the question. \\
Based on these criteria, we conduct the reasoning-level evaluation, requiring integrating \textbf{multiple pieces of visual evidence }from the video context and encouraging questions that can only be solved through \textbf{proactive evidence extraction} from the video, rather than simply disclosing all necessary information in the question. As illustrated in Figure~\ref{fig:examples}, this design systematically and reliably tests models’ abilities for faithful video understanding and complex multimodal reasoning, while tests relying on isolated evidence or fully specified questions often allow shortcut solutions based on single-frame matching or object priors, bypassing the need for genuine video understanding and sufficient reasoning.
}
\vspace{-0.5\baselineskip}

\end{itemize}

\subsection{Construction of \ours}\label{sec:benchmark_construction}
As illustrated in Figure~\ref{fig:construction}, the construction of \ours follows a detailed three-stage pipeline involving task-driven data collection (Sec.~\ref{subsec:data collect}), level-adaptive question-answer annotation (Sec.~\ref{subsec:qa annotation}), and a rigorous multi-stage quality control process (Sec.~\ref{subsec:Quality Control}). Details of the source datasets are provided in Appendix~\ref{sec:appendix-data-source}, and the annotation and review protocols are described in Appendix~\ref{sec:appendix-annotation-pipeline}.
\subsubsection{Task Definition and Data Collection Protocol}
\label{subsec:data collect}
We adopt a task-driven data collection approach, sourcing from diverse curated datasets and internet videos, as illustrated in the top panel of Figure~\ref{fig:construction}. For \LP and \LC, task design and data selection are guided by a systematic survey. Existing human-centric data sources spanning multiple domains and varying scene complexities are reorganized by aligning similar tasks and extracting distinctive aspects as independent tasks. This enables us to curate a diverse set of tasks, each supported by rich and varied datasets.
For \LR, videos are sourced from academic datasets and YouTube videos. The YouTube subset is pre-filtered via domain-relevant tags and then manually reviewed before annotation to guarantee both content richness and safety.

\subsubsection{QA Annotation Protocol}
\label{subsec:qa annotation}
\textbf{\LP and \LC.}
Benefiting from the data collection, we efficiently scale up QA pairs without compromising data quality or diversity by leveraging annotations from existing datasets. 
Task-specific templates and LLM-guided generation are jointly used to create questions and options based on dataset annotations, as illustrated in the lower-left panel of Figure~\ref{fig:construction}.
Moreover, to address underexplored tasks in existing datasets, we manually generate additional QA pairs and complementary annotations by involving domain-specific expert annotators.

\textbf{\LR.}
Experts from each source domain are recruited to annotate questions that encompass five distinct types of reasoning: \textbf{Causal Reasoning}, \textbf{Prediction}, \textbf{Counter-Factual Reasoning}, \textbf{Assessment}, and \textbf{Planning}, under detailed guidelines.
They are asked to create the questions with direct answers and explicitly outline the reasoning steps involved. 
Annotators are also permitted to flexibly select relevant video segments for each question, thereby enhancing contextual diversity and ensuring that the challenges remain grounded in realistic video contexts. 
They can also receive online feedback from the review stage and are given one opportunity to revise their submissions. 
If the revised submission still fails to meet the requirements, the annotation will be discarded.

\subsubsection{Quality Control and Verification}
\label{subsec:Quality Control}
For \textbf{\LP} and \textbf{\LC}, QA pairs are first filtered by LLMs to eliminate those solvable without visual input, followed by human verification conducted by trained annotators. 
Each annotation is carefully reviewed for linguistic quality, answer accuracy, distractor plausibility, and, most importantly, its reliance on visual context.
This pipeline yields over 6,000 high-quality multiple-choice questions.
For \textbf{\LR}, reviewers begin by filling out a detailed checklist that probes each annotation’s objectivity, factual accuracy, non-redundancy, and complexity; they then assign a quantitative score and deliver targeted feedback to the annotator.
Meta-reviewers \textbf{further assess complexity}, ensuring that (1) every question requires integrating multiple visual evidence; (2) the integration pattern cannot be fully determined from the question alone; and (3) each question relies on at least one essential proactive visual evidence. 
The interaction flow among the annotator, reviewer, and meta-reviewer is illustrated in the lower-right panel of Figure~\ref{fig:construction}.
These criteria guarantee \LR’s necessary reasoning challenge and visual-centric novelty.
Common cases of \textbf{valid but insufficiently} complex questions are illustrated in Figure~\ref{fig:examples} upper. This rigorous process yields an acceptance rate below one-fifth, resulting in 442 high-quality questions.
\begin{figure*}[!tbp]
  \centering
    \begin{minipage}[b]{0.32\textwidth} % 改为 [b] 底部对齐
    \centering
    \scriptsize
    {
    \setlength{\tabcolsep}{0.5mm}
    \begin{tabular}{@{}llc@{}}
    \toprule
    & \textbf{Statistics} & \textbf{Value}(Avg./Max) \\
    \midrule
    \multirow{4}{*}[-0.5ex]{Human-P\&C}
    & \# Multi-Choice & 6176 \\
    & \# Options & 4.9 / 5 \\
    & \# Images & 1.1 / 6 \\
    & Video Duration & 35.4 / 584.1 \\
    \midrule
    \multirow{4}{*}[-0.5ex]{Human-R}
    & \# Open-Ended & 442 \\
    & Question Length & 19.2 / 79  \\
    & CoT Length &  86.2 / 183 \\
    & Answer Length & 18.6 / 64 \\
    & Video Duration & 469.3 / 5225.0\\
    \bottomrule
    \end{tabular}
    }
    \captionof{table}{Key statistics of \ours.}
    \label{tab:key_statistics_independent}
  \end{minipage}
  \hfill
  \begin{minipage}[b]{0.26\textwidth} % 改为 [b] 底部对齐
    \centering
    \includegraphics[width=\linewidth]{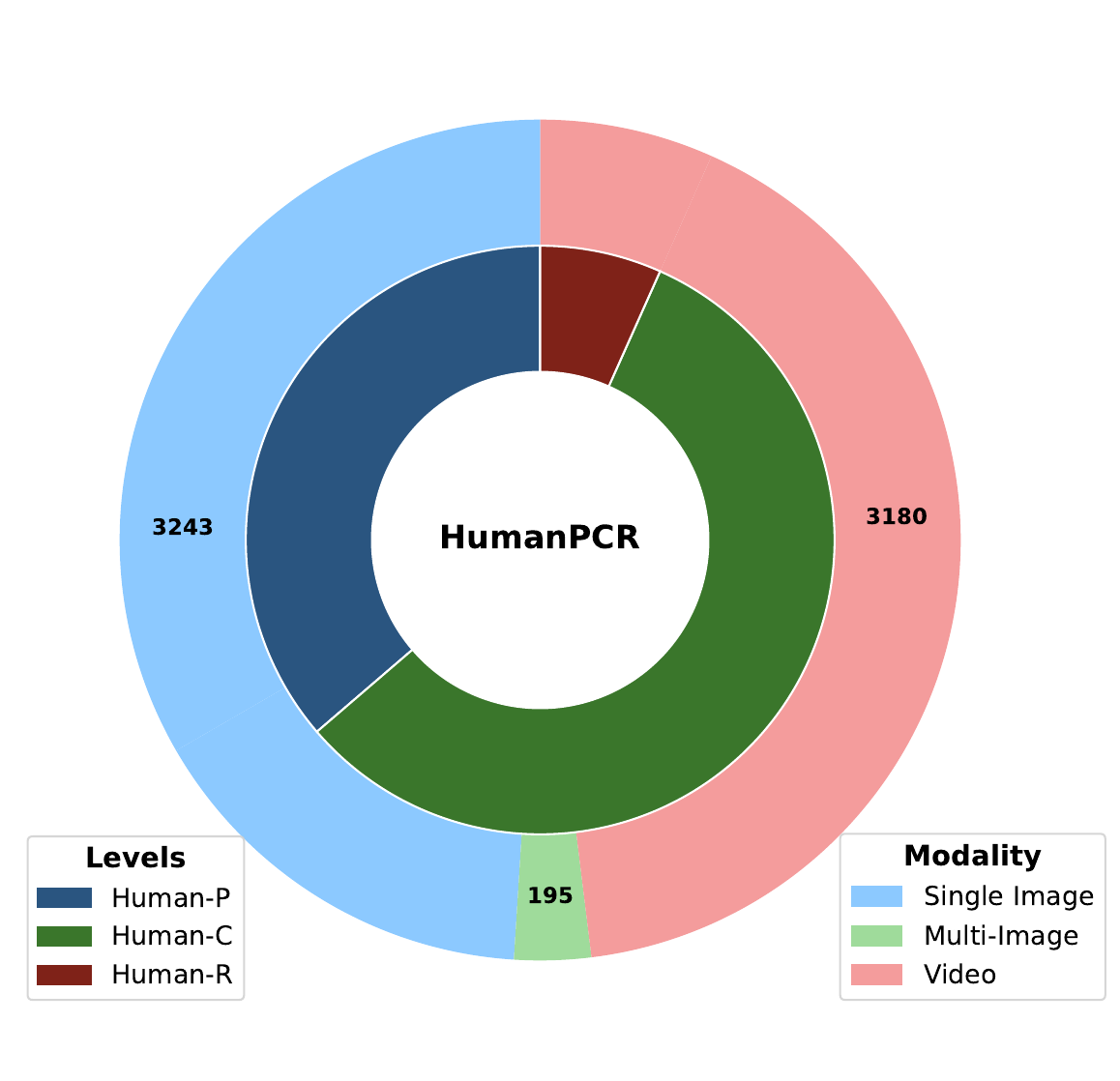} % 使用占位符
    \captionof{figure}{Modality distribution %within each level and modality of 
    in \ours.}
    \label{fig:modality_statistics_independent}
  \end{minipage}
  \hfill
  \begin{minipage}[b]{0.32\textwidth} % 改为 [b] 底部对齐
    \centering
    \includegraphics[width=\linewidth]{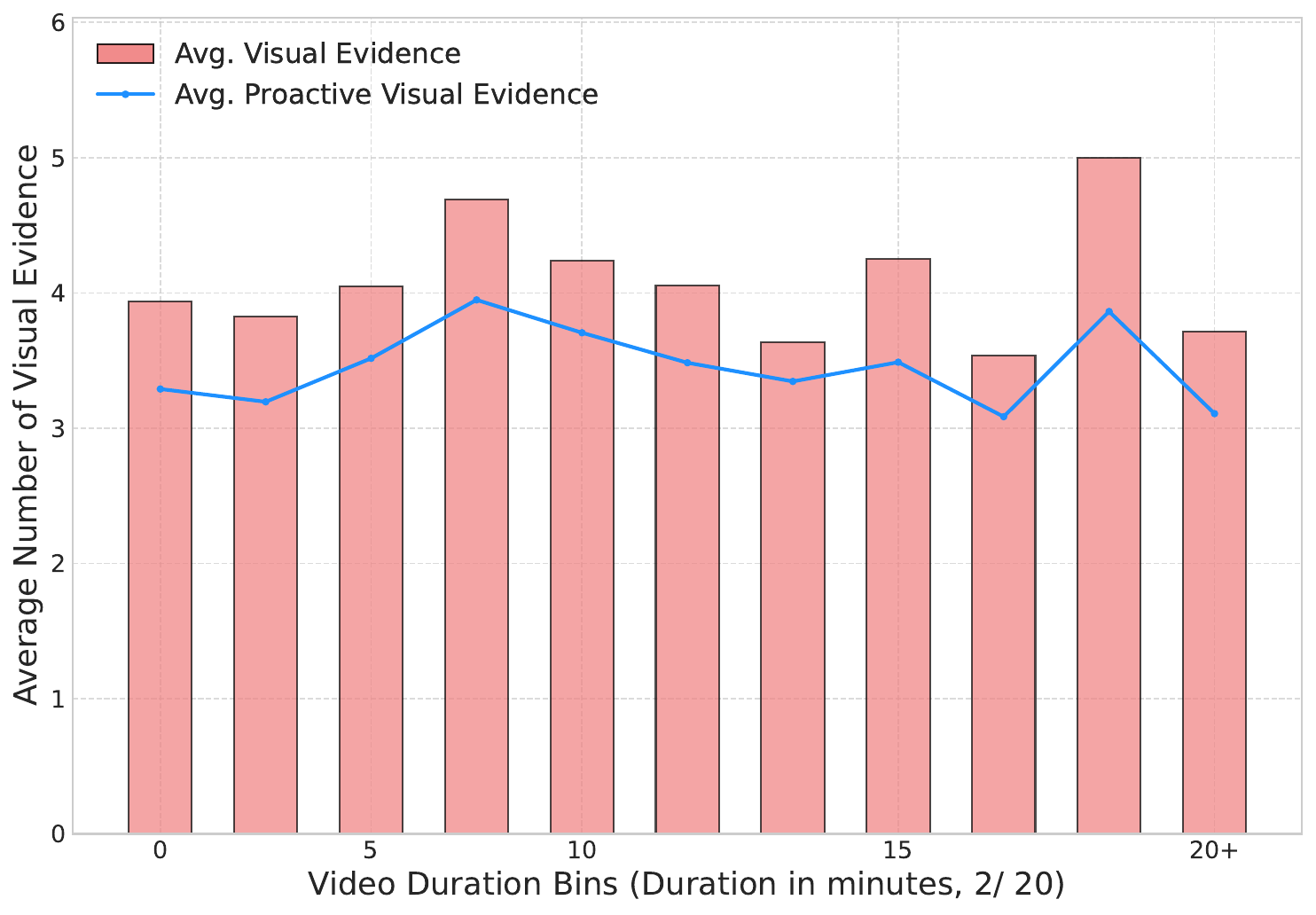} % 使用占位符
    \captionof{figure}{The distribution of the visual evidence number.} %across different duration of \LR.}
    \label{fig:evidence_sta_independent}
  \end{minipage}

  \vspace{-0.3\baselineskip}
  
\end{figure*}
\subsection{Comparison with Existing Benchmarks}
\newcommand{\cmark}{\ding{51}}%
\newcommand{\xmark}{\ding{55}}%

\ours comprises over 6,000 multiple-choice and 442 open-ended questions across hierarchical levels, combining image and video modalities, as presented in Figure~\ref{fig:overview} and~\ref{fig:modality_statistics_independent}. Notably, Figure~\ref{fig:evidence_sta_independent} shows that \LR consistently demands a high level of visual evidence, regardless of video length. This highlights our rigorous quality control procedures and the benchmark’s specialization in visual-centric reasoning, which requires the integration of multiple visual evidence as well as proactive inference.

In comparison to existing benchmarks, \ours fills critical gaps across two domains:
(1) \textbf{Human-centric understanding benchmarks}: Recent works \cite{zhou2024humanvbench}, e.g., MotionBench \cite{motionbench}, ActionAtlas \cite{actionatlas}, and Face-Human-Bench \cite{face_human}, focus on specialized scopes and levles. \ours provides a holistic framework covering a wide range of tasks across diverse dimensions, distinguished by its comprehensive evaluation and multi-faceted task design.
(2) \textbf{Video reasoning benchmarks}: Existing benchmarks fall short of \LR's focus on integrating multiple visual evidence, delving into intricate context, and performing diverse reasoning chains simultaneously.  General video benchmarks~\cite{videomme,zhou2024mlvu,longvideobench} predefine task types with insufficient diversity in visual evidence and reasoning patterns, while automatically-curated works \cite{qi2025vcr, mh-grounded, han2024videoespresso}  often fall short in providing high-quality questions and reasoning processes that are characteristic of human-annotated data. 
Recent benchmarks~\cite{videoholmes,morse500,liu2025videoreasonbench} that emphasize expert knowledge or multi-step reasoning often operate at a much lower level of visual-centric complexity than \LR, as our manual analysis in Figure~\ref{fig:overview} shows: they trigger fewer instances of visual evidence and demand far less proactive evidence seeking. To accommodate complex tasks, these benchmarks rely on video sources with limited scene diversity and real-world complexity—such as short, domain-specific demonstrations~\cite{cao2025videosimple,mmvu}, scripted or program-generated videos~\cite{worldqa,videoholmes,morse500}, or text-heavy presentations~\cite{videommlu,videommmu}. This results in reduced visual evidence diversity and minimal need for evidence integration. In contrast, \LR is built on diverse, human-centric videos, where dense real-world activities and rich implicit rationales naturally support challenging multi-evidence reasoning, without sacrificing authenticity or content variety.

\input{Tables/main_results}

\section{Experiments}

\subsection{Evaluation Settings}
\textbf{Models.} On \ours, we benchmark a diverse set of state-of-the-art MLLMs that support video and image inputs.
Specifically, we evaluate nine proprietary models, among which are Gemini-2.5-Flash\cite{GoogleDeepMind2025Gemini25} and o4-mini\cite{OpenAI2025o3o4mini}.
We also evaluate 30 representative open-source MLLMs, notably including the Qwen-VL series\cite{qwen2_5,qwenvl2} and the InternVL series\cite{internvl,internvl3}.
On \LR, we further include recent methods that advance video understanding, particularly in context extraction and enhanced reasoning\cite{adaretake,video-r1}. Moreover, proprietary models specifically designed for reasoning under strict rate limits are also included, such as o3\cite{OpenAI2025o3o4mini} and Gemini-2.5-Pro\cite{GoogleDeepMind2025Gemini25}.

\textbf{Prompt strategies and frame sampling.} 
Unless otherwise specified, we employ \textbf{Direct Answer} prompts to evaluate models on multi-choice questions and \textbf{Chain-of-Thought (CoT)} prompts for open-ended questions.
For video-based questions, 
we uniformly sample 32 frames for multi-choice questions, while for open-ended questions, we maximize frame utilization by inputting the largest number of frames allowed by each model’s context window and our computational resources. 
Detailed configurations of the evaluated MLLMs are shown in Appendix~\ref{sec:appendix-evaluation-setups}.

\textbf{Metrics.} For multi-choice questions, accuracy is measured by matching the response to the correct option. We report the average accuracy for each task, along with the macro-average accuracy of tasks for each dimension or level. Following prior benchmarks\cite{fang2024mmbenchvideo,mmvu,cao2025videosimple}, we use a quick proprietary model, o3-mini\cite{OpenAI2025o3mini}, as the adjudicator for open-ended questions. 
We evaluate the correctness of responses by comparing the extracted final answers with ground truth.
See Appendix~\ref{sec:appendix-evaluation-setups} for more evaluation details.

\subsection{Main Results}
Table \ref{tab:main-results} presents the main evaluation results of three levels on \ours. Our principal findings are summarized as follows.

\textbf{Current MLLMs still exhibit notable limitations in human-centric visual understanding.} 
As shown in Table \ref{tab:main-results}, tasks in \ours remain largely unsolved by current models. Most models perform below 60\% accuracy on \LP and \LC. Leading models like o4-mini and InternVL3-78B, with overall accuracies of 64.56\% and 63.66\% respectively, still show specific weaknesses, such as \textit{Posture} for InternVL3-78B and \textit{Contact} for o4-mini.  On \LR, o4-mini achieves the highest accuracy of 58.60\%, with most other models below 40\%. These results highlight a substantial challenge in real-world human-centric visual understanding and emphasize the contribution of \ours.

\textbf{Open-source models rival in perception and comprehension, lag in reasoning.}
On \LP and \LC, open-source models match proprietary ones.
Notably, InternVL3-78B surpasses the top proprietary model, o4-mini. 
However, they underperform in reasoning—most remain below 30\% accuracy on \LR, whereas all proprietary models exceed this value. 
Models like o4-mini and Gemini-2.5-Flash illustrate the advantages of proprietary designs for reasoning. 
While open-source models match proprietary ones at perception and comprehension, they struggle on reasoning tasks involving complex visual evidence.

\textbf{Understanding human-centric scenes reflects general capabilities.} 
Considering the results across all dimensions, persistent weaknesses appear on \textit{Posture}, \textit{Contact}, \textit{Behavior}, \textit{Procedure}, and \textbf{reasoning}---all with accuracies around or below 50\%. 
An initial observation is that these results point to deficits in fine-grained perception and spatiality, particularly for recognizing body parts and contact regions under occlusion, consistent with prior findings. 
Poor results on \textit{Behavior} and \textit{Procedure} further expose limitations in temporal understanding and mind modeling, both critical for interpreting human activities and long-term events. 
Overall, the challenges in \ours reveal not only human-centric gaps but also general shortcomings of current MLLMs.

\input{Figures/Analysis/ablation_model_size/figure}

\begin{figure*}[htbp]
  \centering
  \begin{subfigure}[b]{0.63\textwidth}
    \centering
    \includegraphics[width=\textwidth]{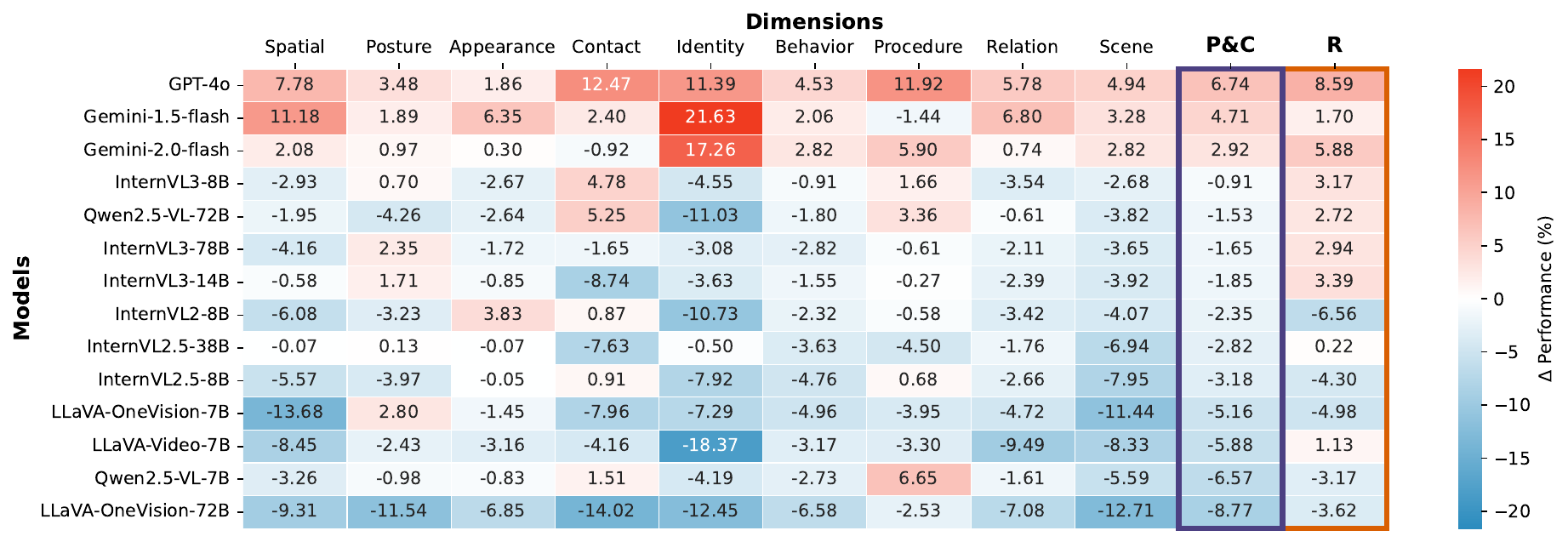}
    \caption{}
    \label{fig:ablation_cot}
  \end{subfigure}
  \hfill
  \begin{subfigure}[b]{0.35\textwidth}
    \centering
    \includegraphics[width=\textwidth]{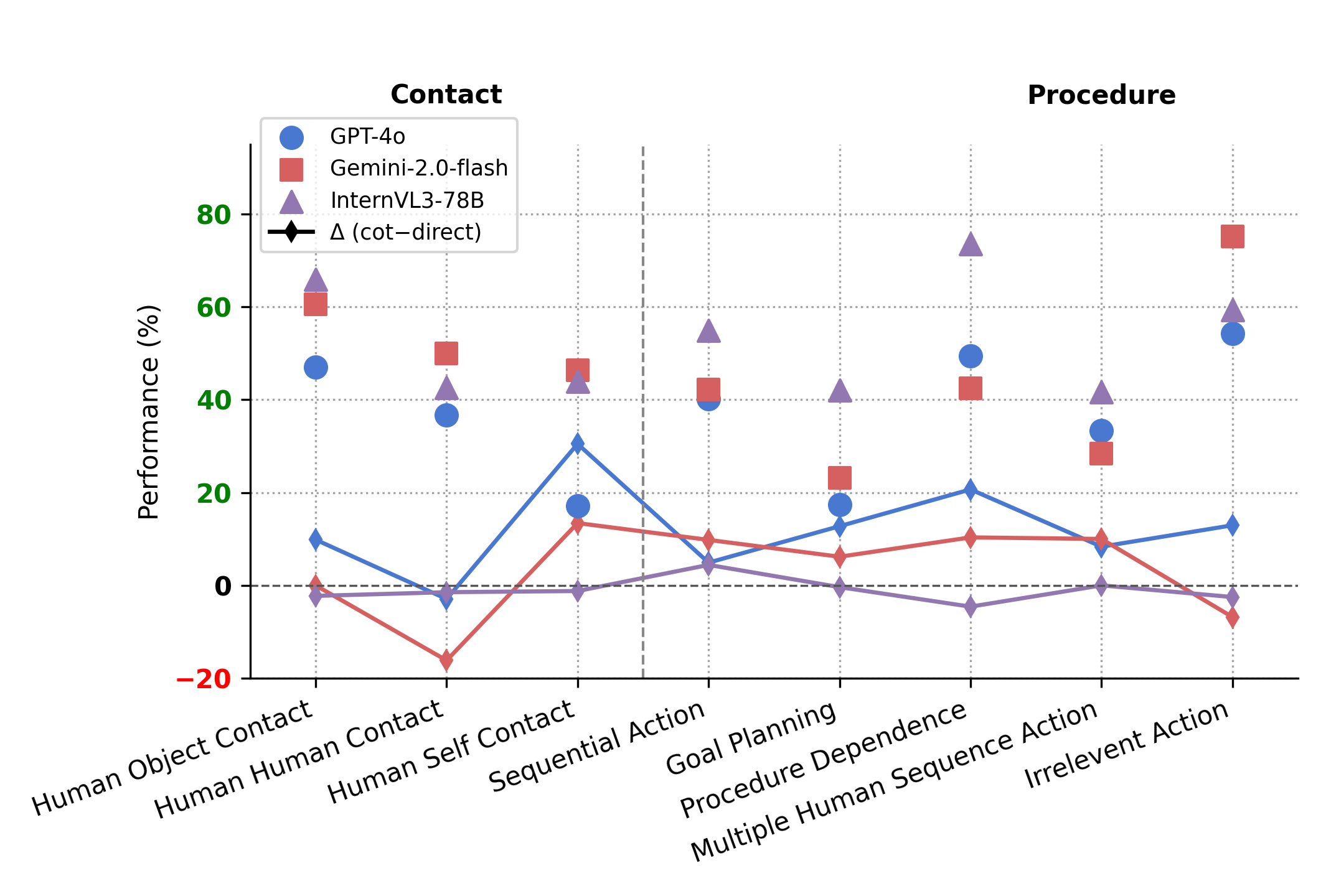}
    \caption{}
    \label{fig:case_study}
  \end{subfigure}
  \caption{Ablation study examining the impact of CoT. 
  (a) Relative improvements of CoT of models on all dimensions and levels.
  (b) Detailed performance across tasks in the \textit{Contact} and \textit{Procedure} dimensions. “$\Delta$ (cot-direct)” represents the accuracy difference between direct answers and CoT-prompted responses.
  }
  \vspace{-0.5\baselineskip}
  \label{fig:twosubs}
\end{figure*}

\begin{figure}[tbp]
  \centering
  \begin{subfigure}[t]{0.24\textwidth}
    \centering
    \includegraphics[width=\linewidth]{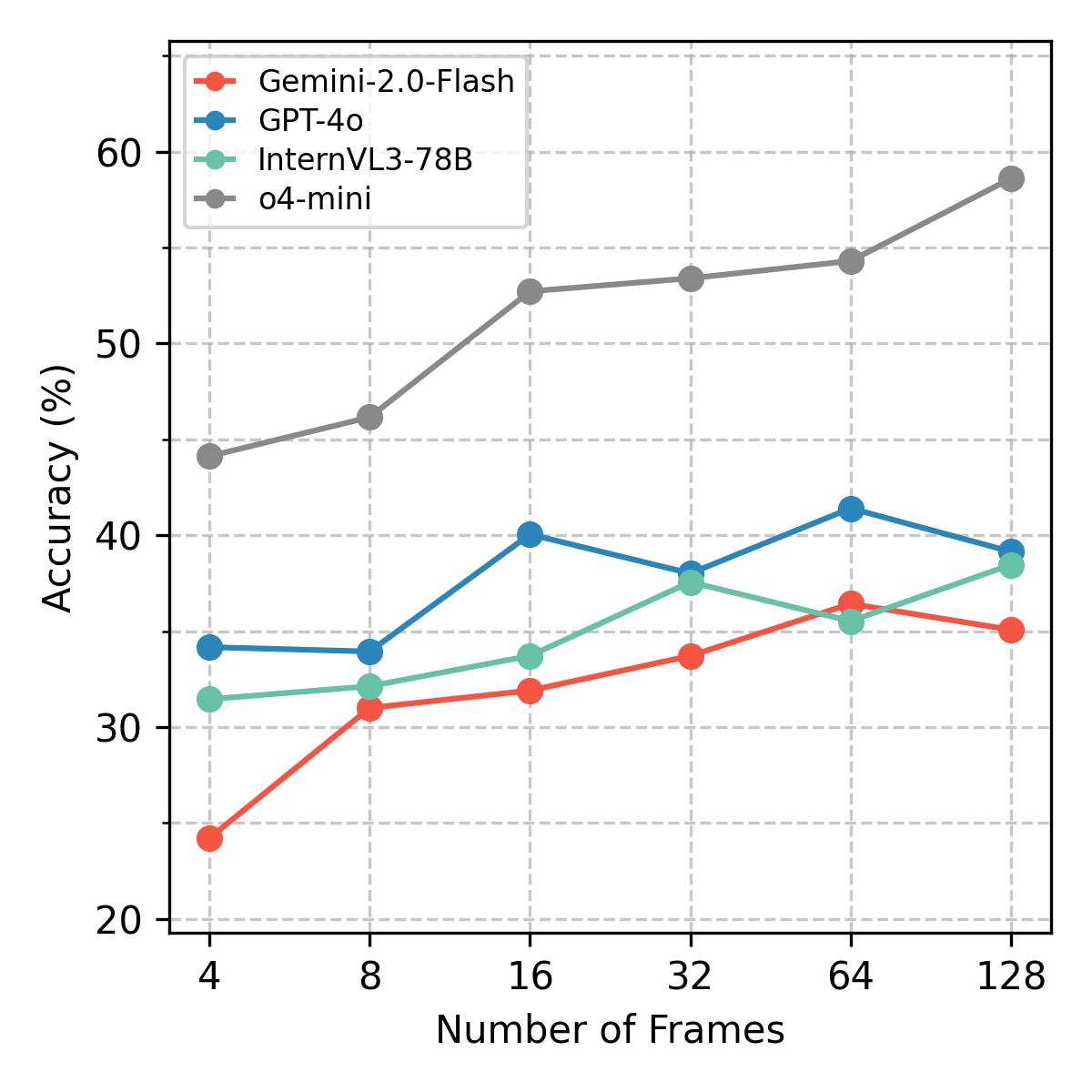}
    \caption{Overall Acc.}
    \label{fig:Overall_Acc}
  \end{subfigure}
  \hfill % 或 \qquad 调整水平间距
  \begin{subfigure}[t]{0.24\textwidth}
    \centering
    \includegraphics[width=\linewidth]{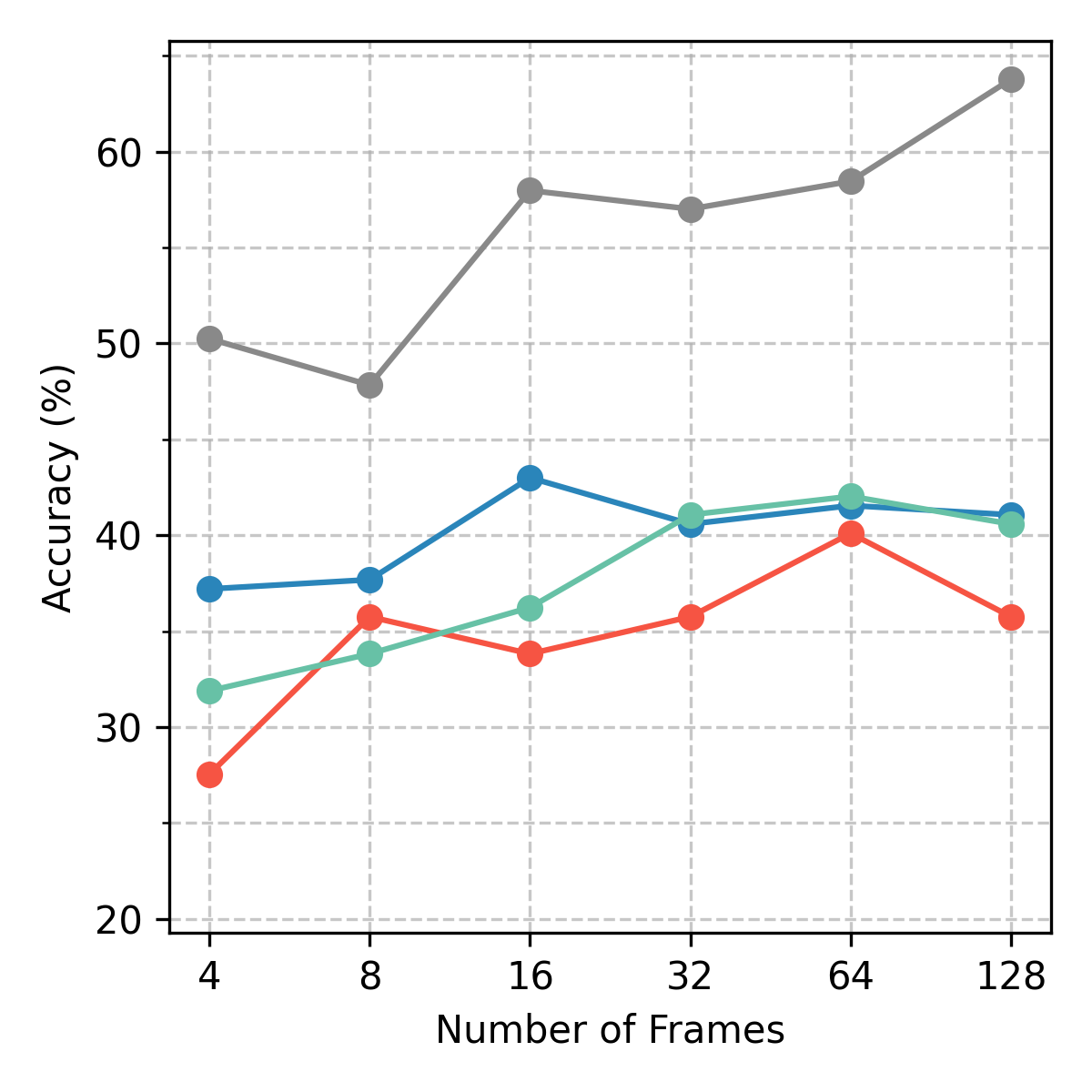}
    \caption{Visual Evidence: 2-3}
    \label{fig:miou}
  \end{subfigure}
  \hfill
  \begin{subfigure}[t]{0.24\textwidth}
    \centering
    \includegraphics[width=\linewidth]{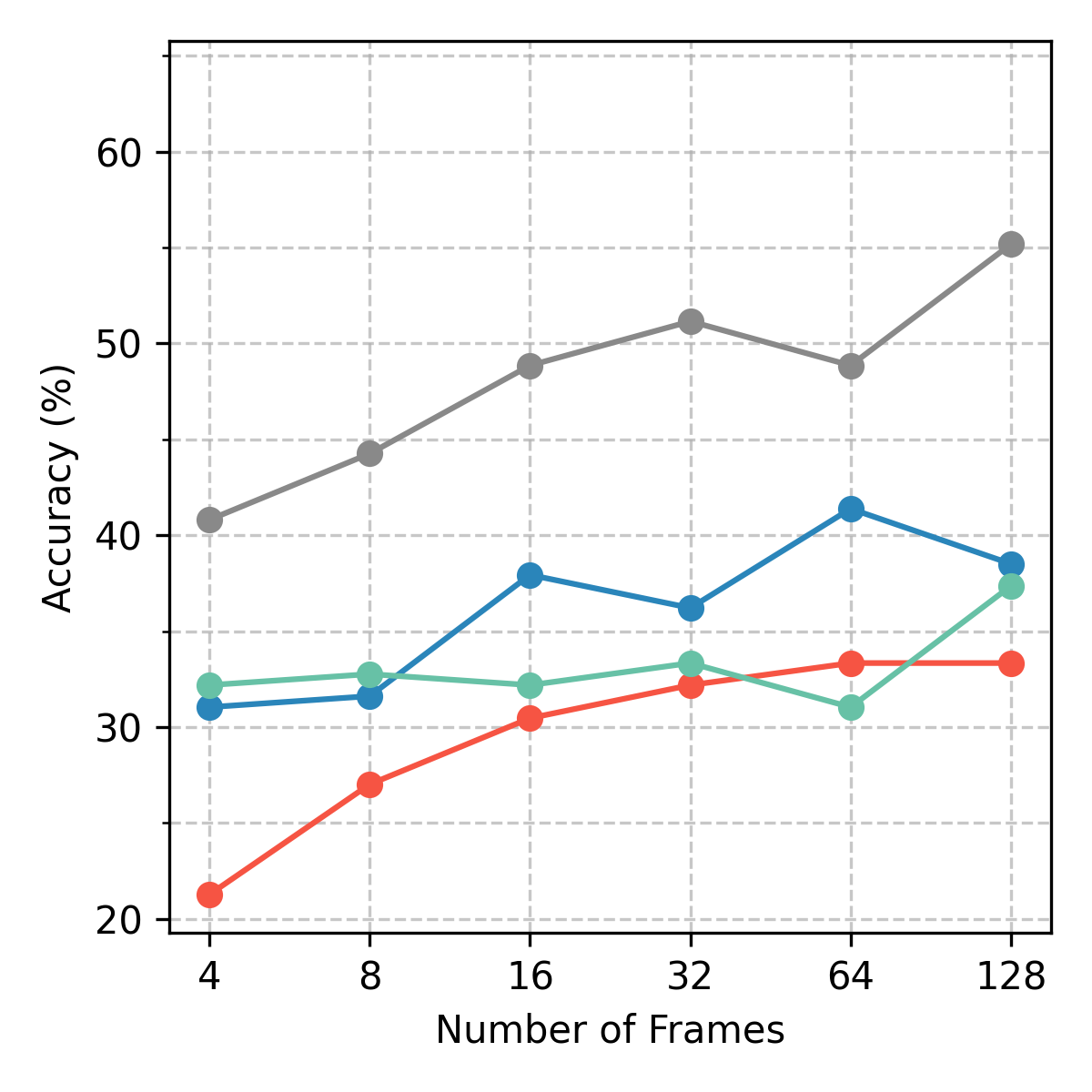}
    \caption{Visual Evidence: 4-5}
    \label{fig:acc-iou}
  \end{subfigure}
  \hfill
  \begin{subfigure}[t]{0.24\textwidth}
    \centering
    \includegraphics[width=\linewidth]{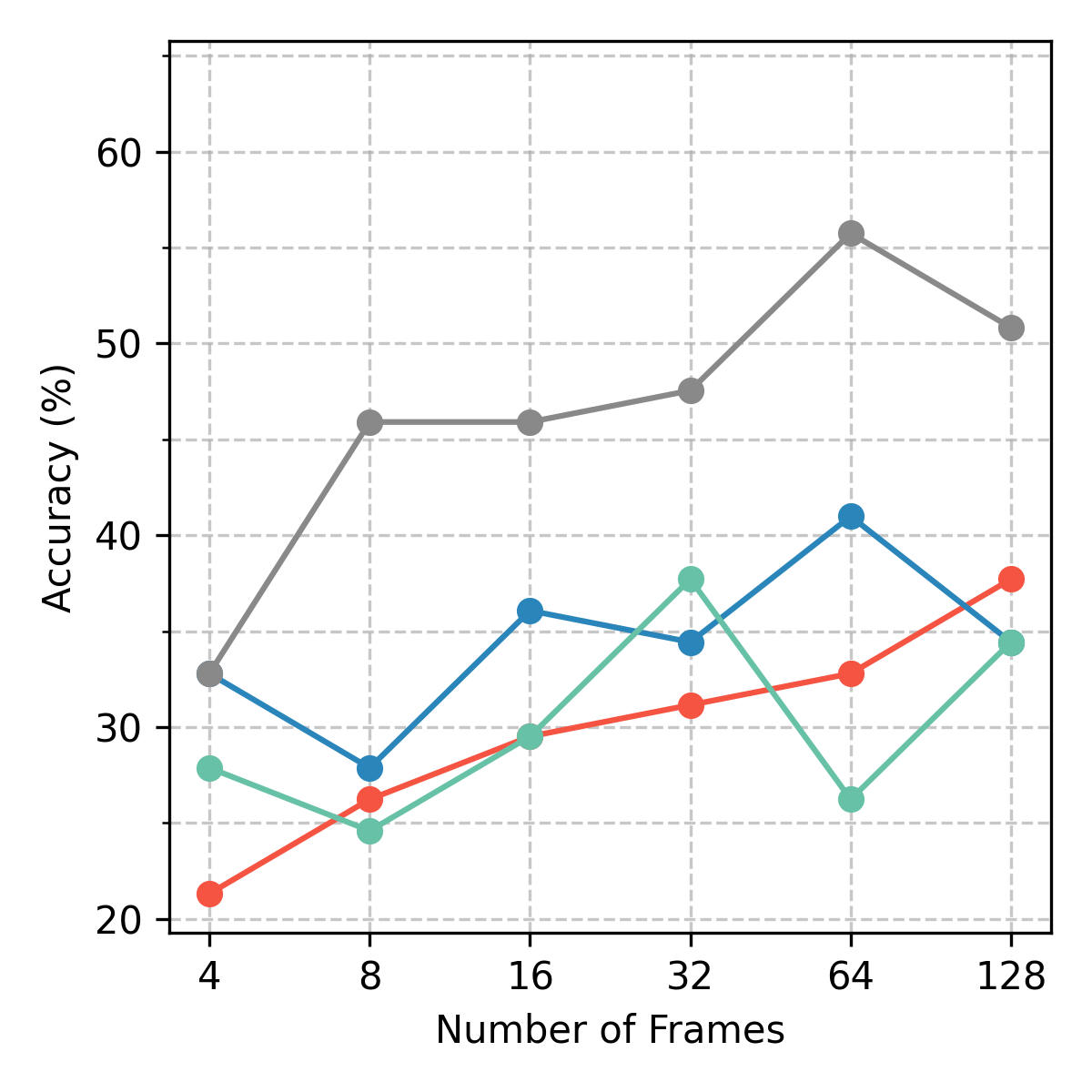}
    \caption{Visual Evidence: 6+}
    \label{fig:crr}
  \end{subfigure}

  \caption{Effect of frame sampling on \LR.
(a) Overall accuracy. (b–d) Accuracy grouped by the number of visual evidence required in the question: (b) 2–3 evidence, (c) 4–5 evidence, (d) 6 or more evidence. Increasing frames shows varying impact depending on visual complexity.}
    \vspace{-0.5\baselineskip}
  \label{fig:frame-num-impact}
\end{figure}

\subsection{Further Analysis} 
To gain a deeper understanding of the capabilities and limitations of frontier models, we conduct a comprehensive analysis focusing on the key factors that influence performance on \ours.

\textbf{Impact of model size.} 
In Table~\ref{tab:main-results}, scaling model size consistently improves accuracy across various model sizes in MLLMs series. 
As shown in Figure~\ref{fig:ablation_model_size}, performance of InternVL2.5 and InternVL3 suggests that complex reasoning tasks are more sensitive to model size scaling, which brings steady improvements, while performance on \LP and \LC plateaus beyond approximately 38B parameters.  
We hypothesize that complex tasks more effectively evaluate the fundamental capabilities of the models, such as world knowledge and logical inference.

\textbf{Mixed CoT effect regarding models and tasks.}
As shown in Figure~\ref{fig:ablation_cot}, the effectiveness of CoT prompting exhibits divergent patterns across models and evaluation dimensions. Specifically, CoT consistently enhances the performance of proprietary models, such as GPT-4o and Gemini-2.0-Flash, whereas it often degrades the performance of open-source models. 
Additionally, the benefits of CoT are dimension-dependent, with improvements more concentrated on complex tasks that demand detailed perception or the integration of multiple visual evidence. 
Figure~\ref{fig:case_study} reveals significant disparities in task-specific accuracy and CoT improvements. This observation highlights the non-redundancy of our taxonomy and emphasizes the necessity of evaluating a diverse set of tasks beyond coarse-grained categories.

\begin{figure}[htbp]
\centering
\includegraphics[width=\linewidth]{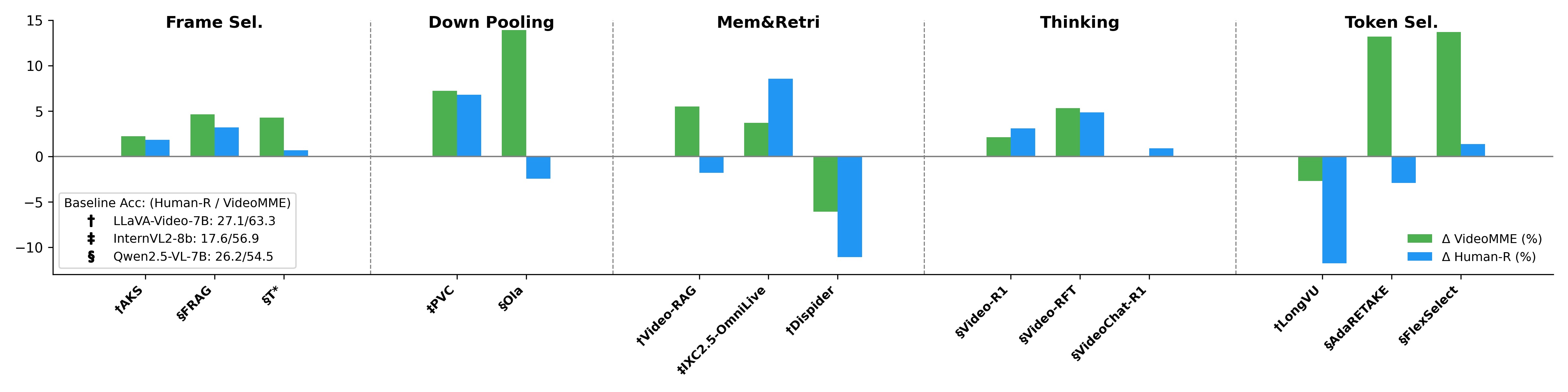}
\caption{Comparisons of advanced video understanding methods on Human-R and Video-MME\cite{videomme}, including RL-tuned \textit{Thinking}\cite{video-r1,videorft,li2025videochat} model and context extraction strategies ranged among \textit{Frame Selection}\cite{aks,tstar,huang2025frag}, \textit{Down Pooling}\cite{yang2024pvc,liu2025ola}, \textit{Memory\&Retrieve}\cite{video-rag, internlm2.5ominilive,qian2025dispider}, and \textit{Token Selection}\cite{adaretake,longvu,zhang2025flexselect}.}
\label{fig:context_extraction}
\end{figure}

\input{Tables/new_test_time_scaling}

\subsection{Delving into Visual Reasoning on Multiple Evidence} 
Based on \LR, we conduct a comprehensive analysis of leading open-source and proprietary MLLMs to gain in-depth insights into the characteristics and limitations of their reasoning capabilities.

\textbf{The impact of frame scaling on multi-evidence reasoning.}  
In Figure~\ref{fig:frame-num-impact}, we examine model performance with varying frame numbers and group results based on the number of visual evidence. Key observations are as follows:
(1) \textbf{Frame number is not the bottleneck.} Scaling the number of frames does not significantly improve reasoning accuracy, indicating that challenges in \LR go beyond vanilla temporal retrieval of evidence.
(2) \textbf{Increased visual evidence exacerbates challenges.} Accuracy decreases as the need for visual evidence increases. For questions requiring 6+ evidence, performance saturates or even worsens with additional frames. We hypothesize that the requirement for multiple visual evidence renders additional context ineffective, while also raising complexity, hindering evidence extraction, and distracting models.

\textbf{Does advanced video understanding configuration help?}
We evaluate MLLMs under advanced video-understanding configurations, context extraction addressing visual redundancy \cite{longvu}
, and test-time compute scaling \cite{cot,deepseek-r1}. Key findings are as follows:
(1) \textbf{Visual Context Extraction.} 
Figure~\ref{fig:context_extraction} compares context extraction methods with uniformly sampled baselines on \LR and Video-MME. 
Validated on Video-MME, most methods achieve only marginal or negative gains on Human-R. In particular, \textit{token selection} and \textit{memory-based retrieval} strategies show clear gaps in performance between the two datasets, while \textit{frame selection} and \textit{pooling} yield consistent improvements. We suggest that token and retrieval methods rely on text-guided token-level filtering, making them prone to missing proactive evidence beyond what the question refers to, unlike the more robust frame-level selection and pooling. This suggests that \LR cannot be fully addressed by query-guided or heuristic methods, posing a new challenge for video reasoning.
(2) \textbf{Test-Time Compute on Reasoning.} Figure~\ref{fig:context_extraction} and Table~\ref{tab:model_comparison_detailed_reward_expanded} show the effect of test-time scaling strategies beyond vanilla CoT. Best-of-N (BoN) delivers over 5\% gains on all 3 models with incremental improvements from stronger reward models or more candidates. Self-Refine merely offers marginal gains and degrades with more iterations. Open-source thinking models Video-R1 improve the baseline by 3.07\%, while proprietary model o3 achieves the best performance of 59.28\%, leaving a large gap to all others. Overall, test-time scaling strategies are generally effective.

\begin{figure*}[htbp]
  \centering
  \begin{subfigure}[b]{0.6\textwidth}
    \centering
    \includegraphics[width=\textwidth]{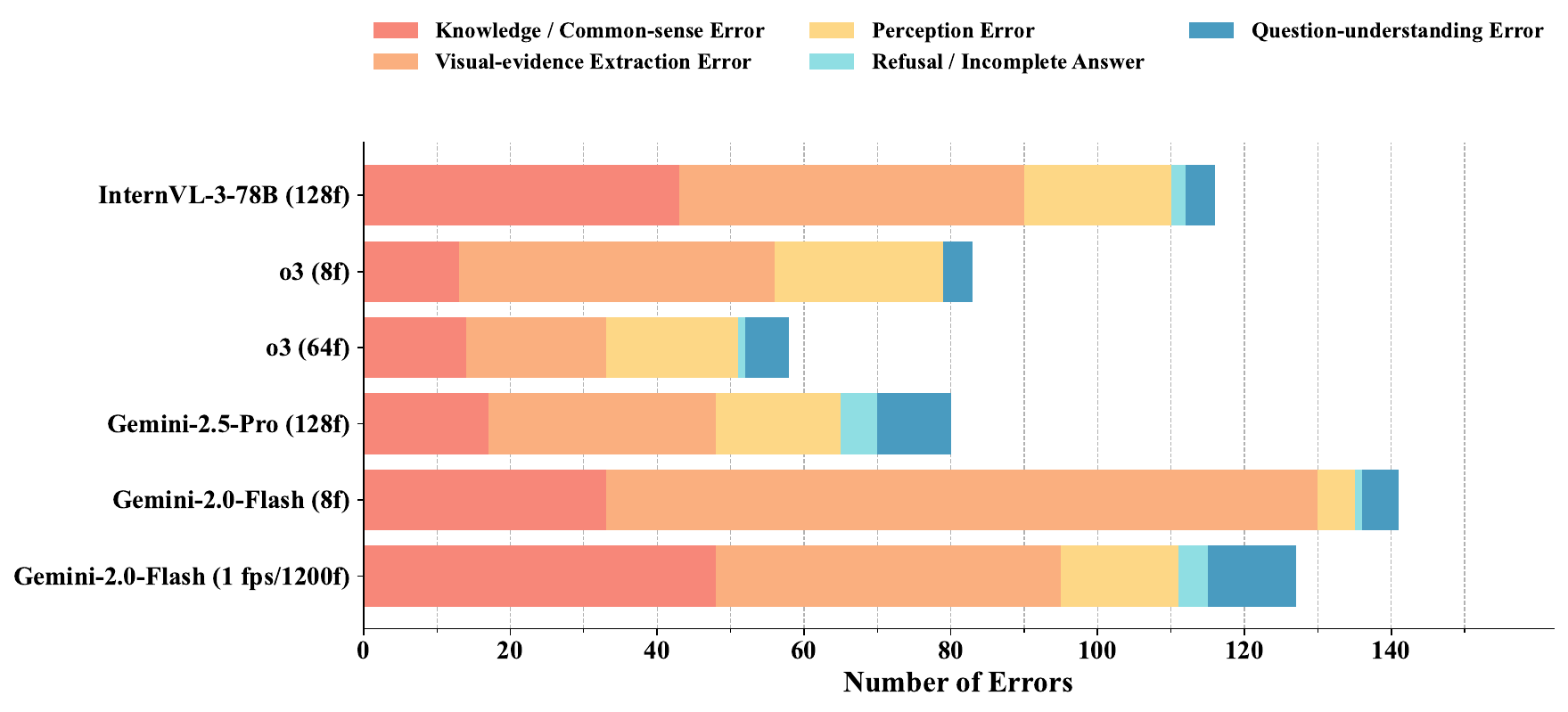}
    \caption{}
    \label{fig:error_length}
  \end{subfigure}
  \hfill
  \begin{subfigure}[b]{0.38\textwidth}
    \centering
    \includegraphics[width=\textwidth]{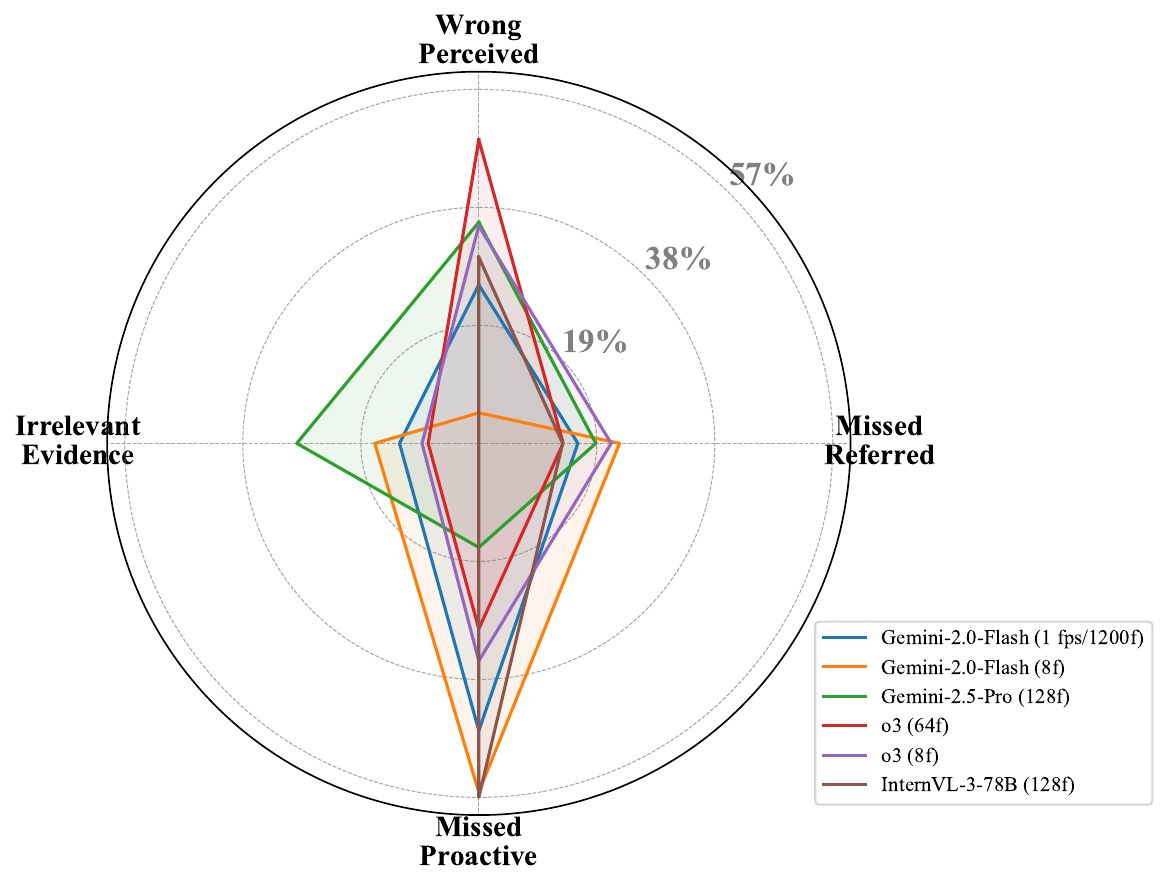}
    \caption{}
    \label{fig:error_radar}
  \end{subfigure}
    \caption{Distribution of error types on \LR across leading models (``f'' means the number of sampled frames).
(a) Counts of five major error types. Most errors fall under visual-related errors --- specifically Perception and Visual-evidence Extraction.
(b) Fine-grained breakdown of visual-related errors. Proactive evidence is more frequently missed than question-referred evidence, suggesting models depend more on question cues than full video context. }
\vspace{-0.5\baselineskip}
  \label{fig:error_bar_plus_radar}
\end{figure*}

\textbf{Error analysis.}
As illustrated in Figure~\ref{fig:error_length}, we analyze 200 randomly sampled questions and classify the errors made by the top-performing models into five categories: Knowledge/Common-sense Error, Visual-evidence Extraction Error, Perception Error, Refusal/Incomplete Answer, and Question‐understanding Error.
Moreover, Visual-evidence Extraction errors are further subdivided into three sub-types. Together with Perception errors, the four visual-related errors are shown in Figure~\ref{fig:error_radar}.
First, Visual-evidence Extraction errors dominate, especially missing proactive evidence not mentioned in the question, which is difficult to resolve even by scaling context (e.g., o3 and Gemini-2.5-Pro with fewer frames outperform Gemini-2.0-Flash with 1200 frames). Second, models heavily rely on query-guided retrieval, as referred evidence in the question is much less likely to be missed, while proactive context understanding remains weak. Third, different models exhibit distinct extraction tendencies, e.g. Gemini-2.5-Pro reduces proactive evidence omissions but introduces more irrelevant evidence compared to o3.
These results suggest that the bottleneck of current MLLMs in video reasoning lies not in factual knowledge gaps but in the inability to holistically model and extract relevant information beyond question cues---a critical aspect that has been largely overlooked in previous evaluations.

\section{Conclusion}
We present HumanPCR, a comprehensive multi-level benchmark designed to rigorously evaluate the capacity of multimodal large language models (MLLMs) to understand humans across diverse real-world scenarios.
It features a systematic, fine-grained taxonomy from hybrid sources and a challenging video reasoning test centered on the integration of multiple visual evidence. \ours reveals persistent challenges, especially in detailed perception, temporal understanding, and complex reasoning. MLLMs often fail to proactively extract key visual evidence in reasoning, relying on query-guided retrieval, with limited gains from scaling visual context or test-time configuration. HumanPCR thus lays a foundation for diagnosing gaps and advancing robust, human-level multimodal understanding. Current limitations of \ours include the reliance on academic datasets for perception and comprehension level, and LLM-based metrics for reasoning. To address these, we plan to extend to more professional domains and develop objective evaluation protocols that retain reasoning difficulty in future work.

{
\small
\bibliographystyle{unsrt}
\bibliography{main}

}

\setcounter{section}{0}
\renewcommand{\thesection}{\Alph{section}}

\refstepcounter{section}
\label{sec:appendix-taxonomy}

\refstepcounter{section}
\label{sec:appendix-data-source}

\refstepcounter{section}
\label{sec:appendix-annotation-pipeline}

\refstepcounter{section}
\label{sec:appendix-evaluation-setups}

\refstepcounter{section}
\label{sec:appendix-data-statistics}

\refstepcounter{section}
\label{sec:appendix-additional-results}

\refstepcounter{section}
\label{sec:appendix-additional-analysis}

\refstepcounter{section}
\label{sec:appendix-data-examples}

\end{document}

%% file: Tables/main_results.tex
{
\begin{table*}[t]
  \centering
  \begin{minipage}{\textwidth}
  
\caption{Results of Proprietary Models and Open-Source Models across three levels of \ours. Macro-average accuracy of tasks within each level and dimension is reported. Full leaderboard is in the Appendix \ref{sec:appendix-additional-results}.}
\label{tab:main-results}
\centering
\scriptsize

\setlength{\tabcolsep}{1.2mm}
\makebox[\textwidth][c]{

\begin{tabular}{lrrrrr>{\columncolor{gray!10}}rrrrr>{\columncolor{gray!10}}r>{\columncolor{blue!10}}r>{\columncolor{green!10}}r}
\toprule
 &  \multicolumn{12}{c}{\textbf{Multi-Choice}} & \multicolumn{1}{c}{\textbf{Open}}\\ 
 \cmidrule(lr){2-13} \cmidrule(lr){14-14}
 
 \textbf{Models} & \multicolumn{6}{c}{\textbf{\LP}} & \multicolumn{5}{c}{\textbf{\LC}} & \multicolumn{1}{c}{\cellcolor{white}} & \multicolumn{1}{c}{\cellcolor{white}\textbf{\tiny{\LR}}} \\

 \cmidrule(lr){2-7} \cmidrule(lr){8-12} \cmidrule(lr){14-14}
 &  {\textbf{Spa.}} &  {\textbf{Pos.}} &  {\textbf{App.}} &  {\textbf{Con.}} &  {\textbf{Ide.}} &  {\textbf{avg.}} &  {\textbf{Beh.}} &  {\textbf{Pro.} }&  {\textbf{Rel.} }&  {\textbf{Sce.}}&  {\textbf{avg.}} &  {\textbf{Acc.}} &  {\textbf{Acc.}}\\

\midrule
Random      & 20.00 & 20.00 & 20.00 & 20.00 & 35.00 & 23.00 & 21.00 & 20.00 & 20.00 & 20.00 & 20.25 & 21.78 & 0.00 \\    

\midrule
\multicolumn{14}{c}{\textbf{Open-source Models}} \\  
\midrule
Aria \cite{li2024aria}  & 79.12  &  39.72  &  61.32  &  36.58  &  76.53  &  55.53  &  48.85  &  41.31  &  50.08  &  55.30  &  48.44  & 51.98   & 28.96  \\
\cmidrule(lr){1-14}
LongVILA-256 \cite{chen2024longvila}   & 75.96  &  31.84  &  63.28  &  38.87  &  41.63  &  49.41   &  44.97  &  31.91  &  51.12  &  53.20  &  44.51 &  46.96   & 21.49  \\
NVILA-8B \cite{liu2024nvila}    & 76.39 & 35.40 & 64.53 & 42.53 & 47.95 & 52.22  & 44.10 & 28.59 & 50.69 & 53.18 & 43.23  & 47.72    & 22.17  \\
\cmidrule(lr){1-14}
MiniCPM-V-2.6 \cite{yao2024minicpm}    & 74.07  &  36.90  &  57.77  &  35.56  &  70.47  &  52.08  &  47.39  &  28.66  &  50.46  &  54.67  &  44.32 &  48.20  & 16.74  \\
MiniCPM-o-2.6 \cite{MiniCPM-o_2.6}  & 80.08  &  43.80  &  63.58  &  36.95  &  71.26  &  56.88  &  47.83  &  31.43  &  53.98  &  55.38  &  46.23  & 51.56   & 21.04  \\
\cmidrule(lr){1-14}
LLaVA-Video-7B \cite{llavavideo}    & 74.73 & 37.13 & 58.03 & 32.62 & 63.50 & 50.99  & 46.18 & 36.26 & 49.54 & 51.77 & 45.37  &  48.18  & 27.60  \\
LLaVA-Video-72B \cite{llavavideo}  & 76.64  &  42.97  &  \underline{71.68}  &  56.54  &  63.63  &  60.49   &  51.74  &  43.73  &  55.41  &  60.92  &  52.41  & 56.45   & 28.28 \\
\cmidrule(lr){1-14}
LLaVA-OneVision-7B \cite{llava_ov}    & 77.95 & 36.33 & 60.03 & 34.13 & 54.63 & 51.02  & 47.67 & 35.56 & 52.40 & 51.06 & 46.02  &  48.52  & 22.85  \\
LLaVA-OneVision-72B \cite{llava_ov}    & 82.57  &  44.77  &  70.40  &  \underline{57.79}  &  71.95  &  62.96   &  51.53  &  43.19  &  58.17  &  61.52  &  52.99 & 57.98   & 27.60 \\
\cmidrule(lr){1-14}
Oryx-1.5-7B \cite{liu2024oryx}   & 74.62  &  36.48  &  62.36  &  42.49  &  39.18  &  50.68   &  44.21  &  34.37  &  51.98  &  49.21  &  44.32  & 47.50   & 22.17  \\
Oryx-1.5-32B \cite{liu2024oryx}   & 82.08  & 40.86  & 64.88  & 47.90  & 45.00  & 55.52  & 47.61  & 44.61  & 54.53  & 57.51  & 50.68 & 53.10   & 28.51  \\  
\cmidrule(lr){1-14}
Qwen2.5-VL-7B \cite{qwen2_5}  & 78.66  & 42.20  & 62.33  & 30.68  & 41.32  & 51.23   & 49.73  & 32.66  & 51.89  & 55.83  & 46.65 & 48.94   & 26.24 \\  
Qwen2.5-VL-72B \cite{qwen2_5}   & 82.11  & \textbf{50.00}  & 68.70  & 43.76  & 41.32  & 57.94  & 55.77  & 46.49  & 53.23  & 64.43  & 54.48 & 56.21  & 34.39  \\
\cmidrule(lr){1-14}
InternVL2.5-8B \cite{internvl2.5}  & 79.43  &  41.68  &  60.37  &  36.69  &  75.45  &  55.83  &  48.53  &  39.13  &  52.24  &  56.47  &  48.51 &  52.17  & 23.53  \\  
InternVL2.5-38B \cite{internvl2.5}   & 84.34  &  45.14  &  68.70  &  50.02  &  84.29  &  63.07   &  53.89  &  \textbf{55.79}  &  54.44  &  63.17  &  56.76 &  59.92  & \underline{35.97}  \\
InternVL2.5-78B \cite{internvl2.5}   & \underline{84.80} & 44.68 & 69.31 & 50.87 & 81.24 & 62.95  & 57.54 & 53.93 & 57.24 & 65.20 & 58.21  & 60.58   & 33.94 \\
\cmidrule(lr){1-14}
InternVL3-8B \cite{internvl3}    & 81.64 & 40.54 & 67.20 & 38.22 & 72.84 & 57.46  & 51.39 & 40.86 & 55.00 & 59.15 & 50.97 &  54.21  & 31.45  \\
InternVL3-38B \cite{internvl}   & 84.73  &  \underline{49.55}  &  69.16  &  \textbf{58.08}  &  \underline{85.89}  &  \textbf{66.15}  &  \underline{57.86}  &  \underline{55.38}  &  \underline{59.01}  &  \underline{66.02}  &  \underline{59.32}&  \underline{62.74}  & 35.75  \\
InternVL3-78B \cite{internvl3}  & \textbf{86.54}  & 46.46  & \textbf{73.39}  & 50.82  & \textbf{86.42}  & \underline{65.34}  & \textbf{57.96}  & 54.31  & \textbf{59.78}  & \textbf{70.21}  & \textbf{60.20} & \textbf{62.77}  & \textbf{37.56} \\
\midrule
\multicolumn{14}{c}{\textbf{Proprietary Models}} \\ 
\midrule
Doubao-1.5-vision-pro \cite{ByteDanceDoubao2024}   & 72.50 & 36.96 & {67.19} & 37.61 & {78.29} & 55.32  & 45.96 & {45.20} & 46.00 & 53.67 & 47.56 &  51.44  & 32.81 \\
Grok-2-Vision \cite{xAI2024Grok2Beta}   & 57.83  & 30.97  & 57.51  & {41.14}  & 50.21  & 46.01   & 46.51  & 42.42  & 42.27  & 56.52  & 46.67  & 46.34  & 36.20  \\
Claude-3.5-Sonnet-v2 \cite{Anthropic2024Claude35Sonnet} & 67.99  & 39.84  & 59.35  & 44.68  & 66.26  & 53.36   & 50.39  & 46.88  & 49.08  & 59.19  & 51.12  &  52.24  & 39.59 \\
Gemini-1.5-Flash \cite{team2024gemini1.5}    & 54.99  & 38.34  & 53.78  & 32.82  & 54.45  & 45.83   & 47.63  & 41.92  & 44.64  & 51.81  & 46.23 &  46.03  & 35.97\\  
Gemini-1.5-Pro \cite{team2024gemini1.5}    & 66.80  &  45.62  &  56.04  &  39.34  &  69.03  &  53.45   &  51.67  &  44.84  &  50.16  &  {61.81}  &  51.69  & 52.57  & 40.05  \\
Gemini-2.0-Flash \cite{GoogleDeepMind2024Gemini2} & {76.42} & {47.46} & 63.75 & \textbf{52.32} & 73.08 & 60.28  & {53.09} & 42.27 & {54.37} & 60.76 & 52.01  &   56.14   & 38.01 \\
Gemini-2.5-Flash \cite{GoogleDeepMind2025Gemini25}& \textbf{82.01}  & \underline{48.10}  & \underline{69.41}  & \underline{49.37}  & \textbf{93.50}  & \textbf{64.66}  & \underline{54.05}  & \underline{49.19}  & \underline{57.27}  & \underline{62.56}  & \underline{55.38} &  \underline{60.02}  & {43.44} \\
GPT-4o \cite{gpt-4o}   & 70.14  &  40.56  &  55.46  &  33.60  &  35.11  &  47.41  &  52.01  &  38.93  &  48.85  &  60.14  &  49.33 &  48.37  & 41.40 \\
o4-mini \cite{OpenAI2025o3o4mini}   & \underline{80.69}  & \textbf{53.13}  & \textbf{71.86}  & 41.09  & \underline{85.89}  & \underline{64.13}  & \textbf{61.10}  & \textbf{54.43}  & \textbf{61.68}  & \textbf{65.97}  & \textbf{60.42}& \textbf{62.28}   & \underline{53.39}  \\  
QVQ-Max~\cite{qvqmax2025} & - & -  & -  & -  & -  & -   & -  & -  & -  & -  & - & -  & 34.84\\  
Claude-3.7-Sonnet\cite{Anthropic2025Claude37Sonnet} & - & -  & -  & -  & -  & -   & -  & -  & -  & -  & - & -  & 40.50\\  
Gemini-2.5-Pro-preview\cite{GoogleDeepMind2025Gemini25} & - & -  & -  & -  & -  & -   & -  & -  & -  & -  & - & -  & 51.13\\  
o3\cite{OpenAI2025o3o4mini} & - & -  & -  & -  & -  & -   & -  & -  & -  & -  & - & -  & \textbf{59.28}\\  

\bottomrule
\end{tabular}%
}
\vspace{-1\baselineskip}
\end{minipage}
\end{table*}

}

%% file: Figures/Analysis/ablation_model_size/figure.tex
\begin{figure*}[t]      % 跨双栏，若不需要跨栏可改为 \begin{figure}[t]
  \centering
  \begin{subfigure}[t]{0.48\textwidth}
    \centering
    \includegraphics[width=\linewidth]{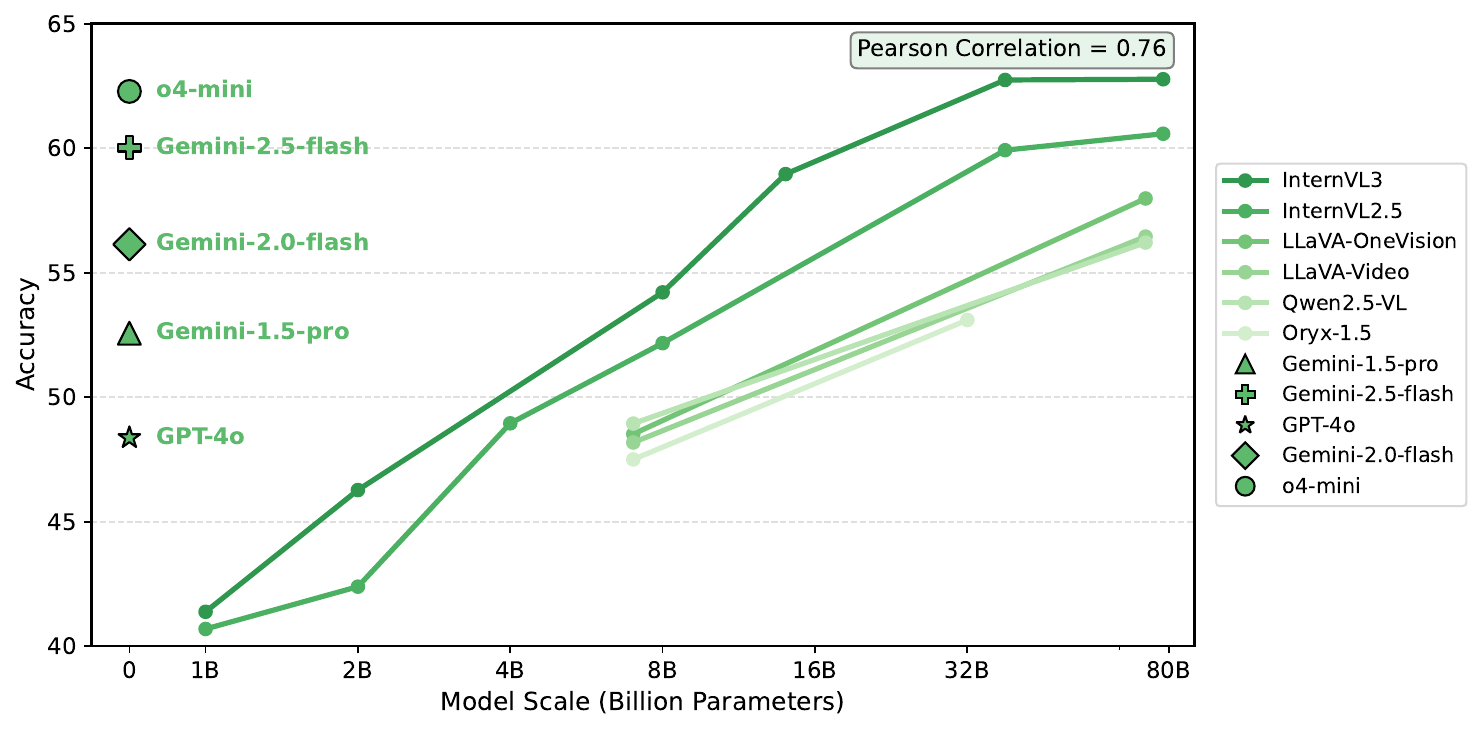}
    \caption{Results on \LP and \LC}
    \label{fig:ablation_model_size_L1L2}
  \end{subfigure}
  \begin{subfigure}[t]{0.48\textwidth}
    \centering
    \includegraphics[width=\linewidth]{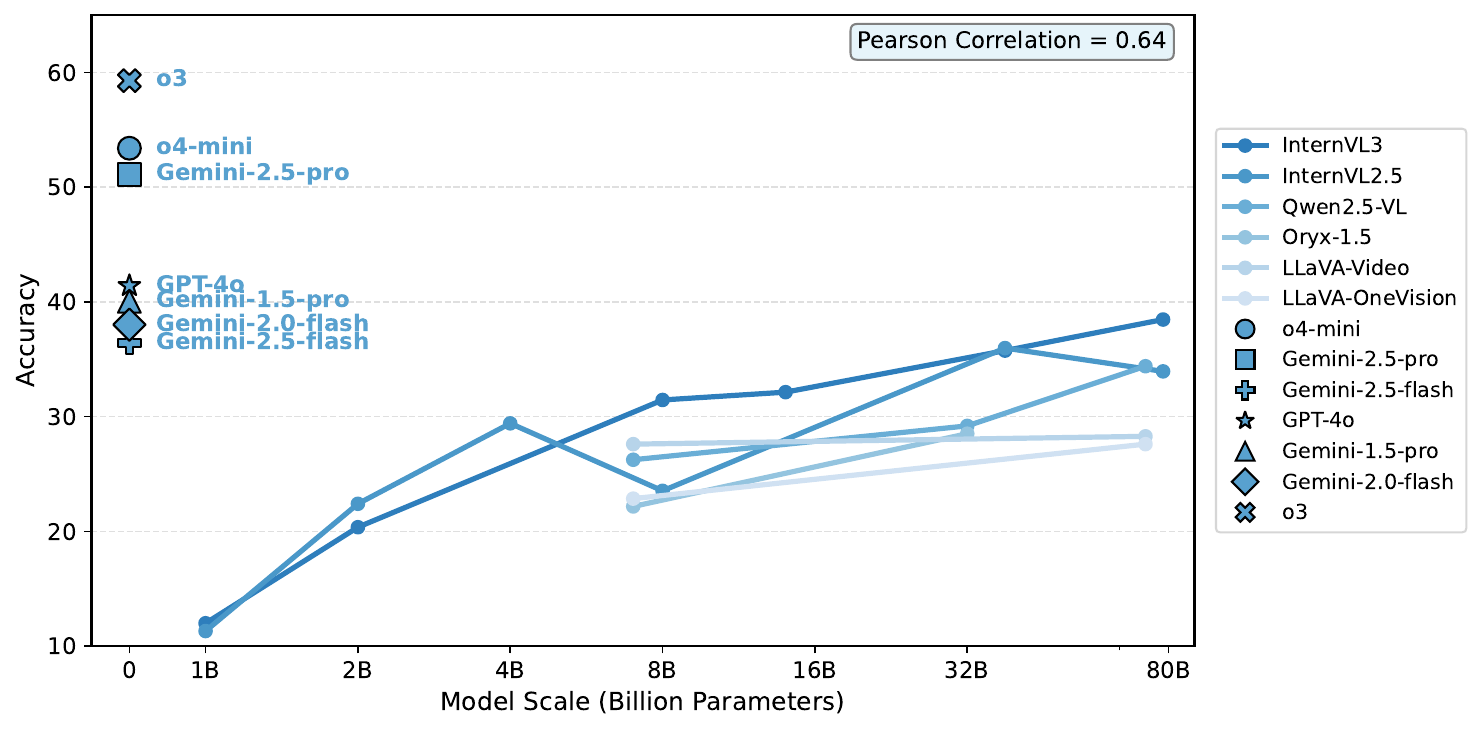}
    \caption{Results on \LR}
    \label{fig:ablation_model_size_L3}
  \end{subfigure}
  \caption{The relationship between model size and macro-average accuracy. }
  \vspace{-0.5\baselineskip}
  \label{fig:ablation_model_size}
\end{figure*}

%% file: Tables/new_test_time_scaling.tex
\definecolor{mygreen}{RGB}{217, 234, 211} % A light green
\definecolor{myblue}{RGB}{208, 224, 243}  % A light blue

\newcommand{\best}[1]{\cellcolor{mygreen}\textbf{#1}}
\newcommand{\secondbest}[1]{\cellcolor{myblue}\underline{#1}}
\newcommand{\formatdelta}[1]{#1}

\begin{table}[htbp]
\centering
\caption{Results of test-time scaling strategies on \LR. Open-source and proprietary thinking models are both included. $M$ represents the magnitude of the strategies. Specifically, the BoN strategy uses $2^M$ candidates, while the Self-Refine strategy performs $M$ iterations.}
\label{tab:model_comparison_detailed_reward_expanded}
\resizebox{\linewidth}{!}{%
\begin{tabular}{@{}lll c c ccc@{}}
\toprule
\textbf{Model} & \textbf{Method} & \textbf{Reward Model} & \textbf{Direct} & \textbf{M=0 (CoT)} & \textbf{M=1} & \textbf{M=2} & \textbf{M=3} \\
\midrule
\midrule
    \multicolumn{8}{c}{{\small\textbf{Test-Time Compute: Prompt Procedure}}} \\
    \midrule
\multirow{4}{*}{\footnotesize Gemini-2.0-Flash}
& BoN    & Gemini-2.0-Flash (Self) & \multirow{4}{*}{32.13} & \multirow{4}{*}{36.43} & 36.88 & 38.01 & 38.69 \\
& BoN    & Gemini-2.5-Flash        &                        &                        & 37.55 & 37.10 & \best{41.86} \\
& BoN    & o4-mini                 &                        &                        & 42.53 & 40.05 & 40.05   \\
& Self-Refine & - &                        &                        & \secondbest{41.40} & 36.65 & 36.65 \\
\midrule
\multirow{4}{*}{\footnotesize GPT-4o}
& BoN    & GPT-4o (Self)           & \multirow{4}{*}{32.81} & \multirow{4}{*}{41.40} & 43.21 & 45.70 & \secondbest{45.93} \\
& BoN    & Gemini-2.5-Flash        &                        &                        & 44.57 & 45.02   & \best{46.38} \\
& BoN    & o4-mini                 &                        &                        & 44.11   & 46.38   & 45.02   \\
& Self-Refine & -    &                        &                        & 39.59 & 39.82 & 40.95 \\
\midrule
\multirow{5}{*}{\footnotesize InternVL3-78B} % Model Name spanning 5 rows
& BoN    & InternVL3-78B (Self)    &  & \multirow{5}{*}{37.56} &32.35&35.06&34.16\\ % 修正了 Direct 列
& BoN    & Gemini-2.5-Flash        &                               &                        & 36.65 & 40.95 & 38.91 \\
& BoN    & o4-mini                 &     \best{35.52}                          &                        & 38.68 & \secondbest{41.40} & \best{42.30} \\ % 从这里移除了 \best{35.52}
& BoN    & GPT-4o                  &                               &                        & 32.8  & 34.6  & 35.29 \\
& Self-Refine & -    &                               &                        & 33.71 & 35.07 & 34.62 \\
\midrule
    \multicolumn{8}{c}{{\small\textbf{Proprietary Thinking Model}}} \\ % 修正了可能的笔误 Property -> Proprietary
\midrule
Gemini-2.5-Pro-preview\cite{GoogleDeepMind2025Gemini25} & - & - & - & 51.13 & \multicolumn{3}{c}{-} \\
Claude-3.7-sonnet\cite{Anthropic2025Claude37Sonnet} & - & - & - & 40.50 & \multicolumn{3}{c}{-} \\
o3\cite{OpenAI2025o3o4mini} & - & - & - & \best{59.28} & \multicolumn{3}{c}{-} \\
\bottomrule
\end{tabular}
}
\end{table}